\documentclass[10pt,twocolumn,letterpaper]{article}

\usepackage{wacv}
\usepackage{times}
\usepackage{epsfig}
\usepackage{graphicx}
\usepackage{amsmath}
\usepackage{amssymb}
\usepackage[accsupp]{axessibility}  %

\usepackage{microtype}
\usepackage{graphicx}
\usepackage{multirow}
\usepackage{booktabs} %
\usepackage[normalem]{ulem}
\usepackage{caption}
\usepackage{subcaption}
\usepackage[dvipsnames]{xcolor}
\usepackage{enumitem}

\usepackage{amsmath, amsthm, amssymb}

\newtheorem{theorem}{Theorem}

\newcommand\boldblue[1]{\textcolor{blue}{\textbf{#1}}}
\newcommand\boldred[1]{\textcolor{red}{\textbf{#1}}}

\def\wacvPaperID{1290} %

\wacvfinalcopy %

\ifwacvfinal
\fi

\ifwacvfinal
\usepackage[breaklinks=true,bookmarks=false]{hyperref}
\else
\usepackage[pagebackref=true,breaklinks=true,colorlinks,bookmarks=false]{hyperref}
\fi

\begin{document}

\title{GraN-GAN: Piecewise Gradient Normalization for Generative Adversarial Networks}

\author{Vineeth S. Bhaskara\thanks{Equal contribution.} $\,^\text{1}$~~~~Tristan Aumentado-Armstrong\footnotemark[1] $\,^\text{1,2,3}$~~~~Allan Jepson$^\text{1}$~~~~Alex Levinshtein$^\text{1}$\\
$^\text{1}$Samsung AI Centre Toronto~~~~~~$^\text{2}$University of Toronto~~~~~~$^\text{3}$Vector Institute for AI\\
{\tt\small \{s.bhaskara,allan.jepson,alex.lev\}@samsung.com,~tristan.a@partner.samsung.com}
}

\maketitle

\ifwacvfinal
\fi

\begin{abstract}
Modern generative adversarial networks (GANs) predominantly use piecewise linear activation functions in discriminators (or critics), including ReLU and LeakyReLU. Such models learn piecewise linear mappings, where each %
piece handles a subset of the input space, and the gradients per subset are piecewise constant. Under such a class of discriminator (or critic) functions, we present Gradient Normalization (GraN), a novel input-dependent normalization method, which guarantees a piecewise $\mathcal{K}$-Lipschitz constraint in the input space. 
In contrast to spectral normalization, GraN does not constrain processing at the individual network layers, and, unlike gradient penalties, strictly enforces a piecewise Lipschitz constraint almost everywhere. 
Empirically, we demonstrate improved image generation performance across multiple datasets (incl.\ CIFAR-10/100, STL-10, LSUN bedrooms, and CelebA), GAN loss functions, and metrics.
Further, we analyze altering the often untuned Lipschitz constant $\mathcal{K}$ in several standard GANs, not only attaining significant performance gains, but also finding connections between $\mathcal{K}$ and training dynamics, particularly in low-gradient loss plateaus, with the common Adam optimizer.

\end{abstract}

\section{Introduction}

\begin{figure}
    \centering
    \begin{subfigure}[b]{0.46\textwidth}
		\includegraphics[width=1.0\textwidth]{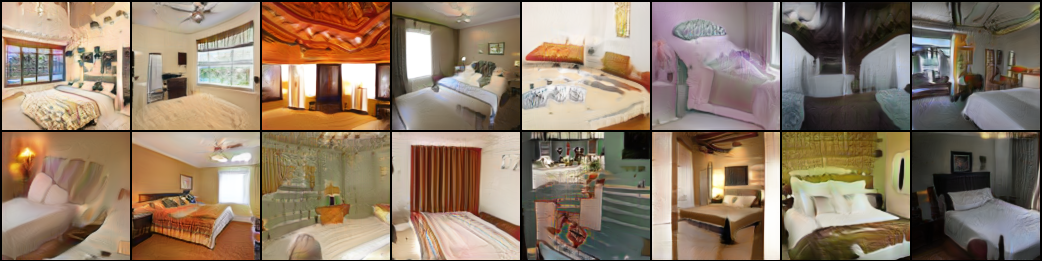}
        \caption{WGAN-GP (FID: 13.6)} 
    \end{subfigure}\\
    \begin{subfigure}[b]{0.46\textwidth}
		\includegraphics[width=1.0\textwidth]{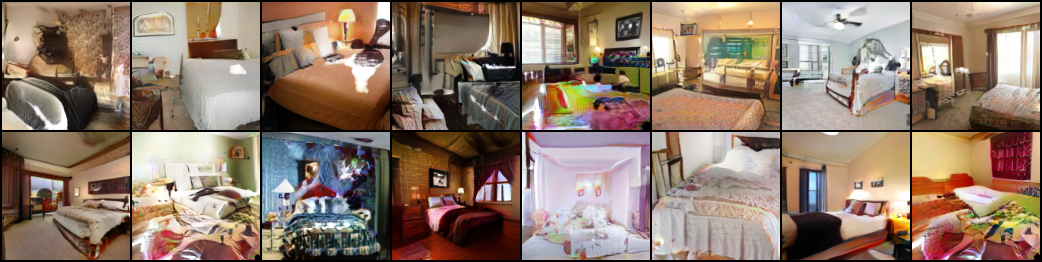}
        \caption{SNGAN (FID: 13.2)}
    \end{subfigure}\\
    \begin{subfigure}[b]{0.46\textwidth}
		\includegraphics[width=1.0\textwidth]{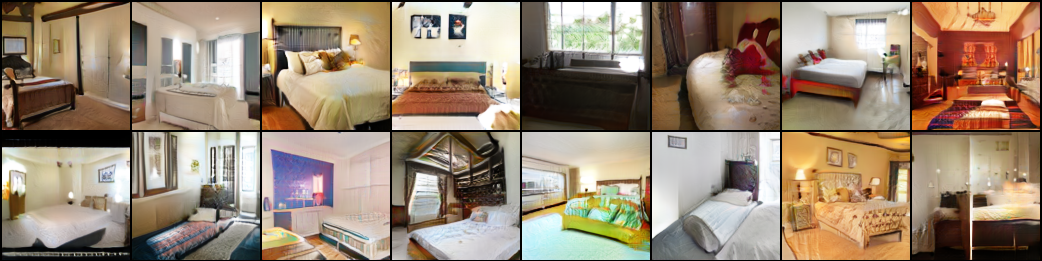}
        \caption{GraND-GAN (FID: 10.8)}
    \end{subfigure}
    \caption{Images generated by WGAN-GP, SNGAN, and our method (GraND-GAN) on $128\times128$ LSUN-Bedrooms (zoom in for better viewing). Lower FID is better.}
    \label{fig:my_label}
\end{figure}

Generative adversarial networks (GANs) \cite{goodfellow2014generative} are a class of generative models that have been shown to be very effective, especially for unsupervised high-resolution image generation \cite{miyato2018spectral,gulrajani2017improved,karras2019style,karras2020analyzing}. GANs usually consist of two networks, a generator $G(z)$ that generates synthetic data conditioned on a noise vector $z$ (sampled from a known noise distribution, 
usually standard normal)
and a discriminator (or critic) $D(x)$ that classifies real data from the generated synthetic data. $G$ and $D$ are generally parametrized as deep neural networks and optimize a mini-max objective. A Nash equilibrium of the zero-sum game is attained when $G$ models the real data distribution and $D$ is maximally uncertain in discriminating the real from synthetic samples.%

Despite the effectiveness of GANs in modeling high-dimensional data distributions, they are hard to train. The quality of the images output by $G$ is dependent on the magnitude of the \textit{input gradients} of the generator loss $\mathcal{L}_G$, written %
\begin{align}
\label{eq1}
\nabla_x\mathcal{L}_G(D(x))= \nabla_{D}\mathcal{L}_G(D(x)) \nabla_xD(x) ,
\end{align}
 where $x=G(z)$. The characteristics of the input gradient of $\mathcal{L}_G$, in general, are determined both by the architecture of the discriminator $D(x)$ as well as the loss function $\mathcal{L}_G$ used in formulating the mini-max objective. In turn, $\nabla_xD(x)$ is a function of the parameters $\theta_D$ of the discriminator (or critic) that is trained to minimize a loss $\mathcal{L}_D$ to separate real samples $x_r$ from the synthetically generated ones $x_f=G(z)$ for a given $G$.
Therefore, in contrast to image classification, the role of $D$ is not only to accurately discriminate real data from fake or synthetic data, but also to have a well-behaved input gradient $\nabla_x D(x)$, which is the primary signal that $G$ relies on for learning. 
Designing such a discriminator is a major objective for GAN research. %

Towards this goal, in this paper, we present a novel input-dependent   normalization method called \textit{Gradient Normalization} or \textit{GraN}, 
which guarantees bounded gradients and a piecewise Lipschitz constraint almost everywhere. GraN can be applied to neural networks with piecewise linear activation functions, a prominent class of function approximators within deep learning, and we use the normalized output for discriminators (or critics). Fig.\ \ref{fig:my_label} shows a qualitative comparison of our method on the popular dataset LSUN-Bedrooms. %

Other works have considered constraining $D$ for stabilizing GAN training. 
In contrast to spectral normalization \cite{miyato2018spectral}, our method does not restrict processing at the individual layers and achieves a tight, piecewise Lipschitz bound on the class of functions that $D$ can model.  
Unlike gradient penalties \cite{gulrajani2017improved}, which softly penalize gradient norms at samples from the input space, our method strictly enforces piecewise constant gradients over the entire input space and ensures the gradient norms within each piece are strictly bounded by a single Lipschitz constant.  However, our normalization introduces discontinuities in the discriminator itself, and it is only piecewise continuous and piecewise Lipschitz.  
Nevertheless,
we show that our method achieves empirical results that are competitive with or better than existing methods.%

Our main contributions are as follows:
\begin{itemize}[itemsep=0pt,partopsep=0pt,topsep=0pt]
    \item We present Gradient Normalization (GraN) for piecewise linear networks $f(x)$ that strictly constrains $f$ to be piecewise $\mathcal{K}$-Lipschitz, where the input space $x$ is partitioned into convex polytopes in each of which $f(x)$ is linear with $\| \nabla_x f(x) \| = \mathcal{K}$.
    \item We show that normalizing discriminators (or critics) with GraN bounds the gradients received by the generator $G$ almost everywhere, stabilizing GAN training.
    \item Unlike spectral normalization (SN), GraN does not restrict processing at the individual layers and does not suffer from gradient attenuation. Further, in contrast to both SN and gradient penalties, GraN enforces the piecewise Lipschitz property as a hard constraint. 
    \item Empirically, GraN performs better or competitive to existing methods on multiple datasets (CIFAR-10, CIFAR-100, STL-10, LSUN bedrooms, and CelebA), and two loss functions (discriminators with a non-saturating cross-entropy loss and critics with a soft hinge loss).
    \item While GraN only enforces a local, piecewise Lipschitz constraint, we find the finite-difference gradient norm is empirically well-behaved across large step sizes, likely including jumps across the polytopes modeled by $f$. 
    \item We also investigate the effect of the Lipschitz constant $\mathcal{K}$ on standard baseline models, finding (a) constrained discriminators (trained with cross-entropy loss) outperform constrained critics, (b) $\mathcal{K}$ influences training dynamics when using Adam, especially on loss plateaus, and (c) tuning $\mathcal{K}$ can significantly improve performance.
\end{itemize}

\section{Related Work}
\label{relwork}
\subsection{Stabilizing GANs}
\label{relwork:stab}

Most previous works on stabilizing GANs take one of the following approaches:
(1) proposing novel loss functions 
  (e.g., \cite{mao2017least,berthelot2017began}),
(2) devising improved architectures
  (e.g., \cite{karras2020analyzing,zhang2019self,zhao2020diffaugment,park2021styleformer,gong2019autogan,ghosh2018multi,karras2017progressive})
or
(3) introducing new constraints on $D$
  (e.g., \cite{arjovsky2017towards,arjovsky2017wasserstein,roth2017stabilizing,tseng2021regularizing}).
Herein, we continue along the latter line of research, constructing an architectural regularization on $D$ that improves training without sacrificing network capacity. 

One of the earliest works on constraining $D$ to stabilize GAN training is the Wasserstein GAN (WGAN) \cite{arjovsky2017wasserstein}. They propose a novel loss function and present weight-clipping as a way to regularize $D(x)$ to be 1-Lipschitz in $x$. However, subsequent work \cite{gulrajani2017improved} showed gradient penalties to be more effective, as they do not impede optimization or severely reduce network capacity.

\subsection{Gradient Penalties}
\label{relwork:gradpen}

Recent research has found gradient penalties (GPs), 
of the form
$ %
P_\delta(x) = ( || \nabla_x f(x) ||_2 - \delta )^2,
$ %
to be useful for GANs.
Building on WGAN, 
among the most popular GAN variants is WGAN-GP \cite{gulrajani2017improved}, which forgoes weight clipping by applying $P_1$ along random convex combinations of reals and fakes, since any $C^0$ function with unit-length gradient is necessarily 1-Lipschitz. 
Later research considered \textit{one-sided} GPs 
\cite{petzka2017regularization}, as well as alternate sampling methods \cite{wei2018improving}.

Part of the motivation for turning to optimal transport distances is the reduction of gradient uninformativeness, which is a problem for most $f$-divergence-based GANs when the real and fake distributions do not sufficiently overlap \cite{roth2017stabilizing}.
However, Zhou \etal \cite{zhou2019lipschitz} show that Lipschitz continuity can combat gradient uninformativeness in GANs more generally. 
Separately, Roth \etal \cite{roth2017stabilizing} showed that increasing distributional overlap via noise is approximately equivalent to a zero-centered GP.
This was simplified to the popular ``R1'' GP (defined as $P_0$ on real data) \cite{mescheder2018training}, used in recent state-of-the-art GANs 
(e.g., \cite{karras2020analyzing,chan2020pi}). 

Clearly, gradient regularization has seen empirical success  in improving GANs.
Yet, a downside of soft GPs is that they may not enforce exactly the desired value at a given position; furthermore, they are only applied to a subset of the input domain, which may shift over time.
In contrast, GraN enforces unit gradients almost everywhere by construction.

\subsection{Weight Normalization}

Weight normalization (WN) \cite{salimans2016weight}  reparametrizes layers in a manner conducive to better conditioned optimization,  used in early GAN work \cite{salimans2016improved}.
In WN, for each individual layer $i$ of the network, the corresponding weight vectors are rewritten as
$%
\widetilde{w_i} =  w_i {\rho_i} / {|| w_i ||}~\forall~i,
$ %
where the learned scalar $\rho_i$ controls the norm and $w_i/{|| w_i ||}$ represents the direction.

For piecewise linear networks, 
each $x$ is linearly mapped to 
$f(x) = w\cdot x + b$, where $w$ is locally constant around $x$, 
is implicitly  input-dependent, 
and may be interpreted as the \textit{effective weight vector} of a (local) linear model 
(see \S\ref{sec-pieces} for details).
In a manner reminiscent of WN, 
GraN essentially normalizes $f$ by the norm of its gradient, i.e., $\nabla_xf=||w||$.
In contrast, however, GraN acts on the full network, rather than a single layer at a time, and enforces piecewise constant gradients, rather than reparametrizing the network.

\subsection{Spectral Normalization and Gradient Attenuation under Global Lipschitz Constraints}

Building on prior regularizations, 
such as GPs and weight clipping,
Miyato \etal \cite{miyato2018spectral} present spectral normalization (SN) as an alternative method of ensuring $D$ is 1-Lipschitz, 
without an additional penalty in the objective for $D$. 
This is enforced by a layer-wise weight normalization technique,
dividing by an estimate of the maximal singular value from each weight matrix.
Empirically, SN is an effective stabilizer for GANs during training, 
independent of the loss function employed.
As a result, it is a major component in recent large-scale GANs \cite{brock2018large,schonfeld2020u}. 
We discuss SN further in \S\ref{sec-comparisons}.

At an architectural level, 
balancing network capacity and regularization is difficult for globally Lipschitz-constrained networks.
Indeed, Anil \etal \cite{anil2019sorting} showed that standard networks struggle to solve simple tasks when globally Lipschitz constrained, and that smooth SNed ReLU networks with unit gradients become globally linear, 
which they solve by enforcing gradient norm preservation and weight matrix orthonormality.
We avoid this in GraN by permitting discontinuities in $D$ with respect to the input space and by only constraining the Lipschitz constant $\mathcal{K}$ locally in a piecewise manner.
Later work \cite{li2019preventing} 
combated the gradient attenuation induced by Lipschitz constraints 
with a novel orthogonal convolution operator.
In contrast, %
GraN
can be applied on top of any piecewise linear network without globally constraining $\mathcal{K}$, while doing so locally in a piecewise sense.

\section{Background}

\subsection{Generative adversarial networks}
\label{sec-background-gans}
Let $\mathbb{P}_r$ represent the distribution of real data and  $\mathbb{P}_g$ be the distribution of generated data at a given state of the generator $G$. Let $\mathcal{L}_G$ and $\mathcal{L}_D$ represent the loss functions for the generator $G$ and the discriminator (or critic) $D$, respectively. Let $z\sim \mathcal{N}(0, 1)$ be a $|\mathcal{Z}|$-dimensional noise vector sampled from an i.i.d.\  standard normal distribution. Let $f:\mathbb{R}^{d} \rightarrow \mathbb{R}$ represent a deep neural network encoding a scalar field.
For image generation, 
$ \mathbb{R}^{d} = \mathbb{R}^{3 \times H \times W} $.

Goodfellow \etal
\cite{goodfellow2014generative} originally propose a cross-entropy (CE) loss-based objective for training $G$ and $D$ as follows:
\begin{align}
    &\mathcal{L}_D = \mathbb{E}_{x \sim \mathbb{P}_r}\left[ -\log D(x) \right] + \mathbb{E}_{x \sim \mathbb{P}_g}\left[ -\log (1-D(x)) \right],\label{eqD1} \\
    &\mathcal{L}_G = \mathbb{E}_{x \sim \mathbb{P}_g}\left[\log (1-D(x)) \right]\label{eqG1},
\end{align}
where the discriminator $D(x) = \sigma(f(x))\in [0, 1]$ represents the probability of a sample $x$ coming from the real distribution $\mathbb{P}_r$ and $\sigma(\cdot)$ is the sigmoid function $1/(1+e^{-(\cdot)})$. 
Hence, $f$ computes a logit mapped to a probability by $\sigma(\cdot)$. 
The gradient of $\mathcal{L}_G$ with respect to the inputs $x$ 
is then
\begin{align}
\nabla_x\mathcal{L}_G = \mathbb{E}_{x\sim \mathbb{P}_g}\left[-D(x)\nabla_xf(x)\right].
\end{align}
Notice that, 
when $D(x)\rightarrow 0$ for some generated $ x \sim \mathbb{P}_g$,
the contribution to $ \nabla_x\mathcal{L}_G $ from such points will necessarily be small and have little influence on updating $G$.
This is particularly probable early in training,
when $\mathbb{P}_g$ and $\mathbb{P}_r$ are easily separated.
To overcome this problem, in the same work, the authors suggest using an alternative objective for $G$ that is non-saturating, given as
\begin{align}
    \mathcal{L}_G = \mathbb{E}_{x \sim \mathbb{P}_g}\left[-\log D(x) \right]\label{eqG2},
\end{align}
for which the gradient is
\begin{align}
\nabla_x\mathcal{L}_G=\mathbb{E}_{x\sim \mathbb{P}_g}\left[-(1-D(x))\nabla_xf(x)\right]. \label{vanillagangrad}
\end{align}
In this case, note that when $D$ confidently rejects fakes  
(i.e., $D(G(z)) \approx 0$),
they can still contribute to $\nabla_x\mathcal{L}_G$.
For future convenience of notation, we use ``NSGAN" to refer to the non-saturating (NS) version of training a discriminator via a GAN, where $\mathcal{L}_D$ is given by Eq. \eqref{eqD1} and $\mathcal{L}_G$ is given by Eq. \eqref{eqG2}. Note that although $\mathcal{L}_G$ in Eq. \eqref{eqG2} is relatively non-saturating early in the training compared to Eq.\  \eqref{eqG1}, it may still saturate later in training as $\mathbb{P}_g$ gets closer to $\mathbb{P}_r$.

Subsequently, the WGAN model \cite{arjovsky2017wasserstein} attempted to address two issues with the original GAN formulation. First, due to the use of the Jensen-Shannon divergence, disjoint support for $\mathbb{P}_g$ and $\mathbb{P}_r$ leads to poor gradients (as a confident $D$ rapidly saturates, leading to vanishing gradients). Second, for fixed $D$, the optimal $G$ samples a sum of Dirac deltas at points with the \textit{highest} value of $D(x)$, leading to gradients that encourage mode collapse \cite{metz2017}. 
Using an approximate Wasserstein distance in WGANs mitigates these issues.

This WGAN objective for $G$ and critic  $D$ may be written 
\begin{align}
    &\mathcal{L}_D = -\mathbb{E}_{x\sim\mathbb{P}_r}\left[ D(x) \right] + \mathbb{E}_{x\sim\mathbb{P}_g}\left[ D(x) \right]\label{wgand1},\\
    &\mathcal{L}_G = - \mathbb{E}_{x\sim\mathbb{P}_g}\left[ D(x) \right]\label{wgang1},
\end{align}
where $D(x)=f(x)$, and $f(x)$ is constrained to be 1-Lipschitz in $x$, denoted by $\|f\|_{\text{Lip}}=1$. We now have
\begin{align}
    \nabla_x\mathcal{L}_G=\mathbb{E}_{x\sim \mathbb{P}_g}\left[-\nabla_xf(x)\right], \label{wgangrad}
\end{align}
which does not saturate, unlike NSGAN 
(Eq.\ \eqref{vanillagangrad}). 

Another commonly used variant of the critic loss $\mathcal{L}_D$ 
is the hinge loss \cite{lim2017geometric,miyato2018spectral}:
\begin{align}
    \mathcal{L}_D = \mathbb{E}_{x\sim\mathbb{P}_r}&\left[ \text{ReLU}\left(1-D(x)\right) \right] +\nonumber \\&\mathbb{E}_{x\sim\mathbb{P}_g}\left[ \text{ReLU}\left(1 + D(x) \right) \right]\label{wgand2},
\end{align}
where $\text{ReLU}(x)=\max\{0, x\}$ is the rectified linear unit. 
It is straightforward to define a smooth analogue of the hinge loss 
in Eq.\ \eqref{wgand2}
by using a smooth approximation of ReLU defined as $ \text{SoftPlus}(x) = \log(1+e^x) $,
which we refer to as the \textit{soft hinge loss}. For convenience of notation, we call $D$ a \textit{critic} with $D(x)=f(x)$, when $\mathcal{L}_D$ is any of the Wasserstein, hinge, or soft hinge loss, and a \textit{discriminator} with $D(x)=\sigma(f(x))$, when $\mathcal{L}_D$ is a variant of the cross-entropy loss (as in NSGANs).

For GANs,
several methods have been proposed to effectively constrain $D$, generally to 1-Lipschitz function spaces. 
WGAN uses weight clipping, which compromises optimization ease and capacity,
while WGAN-GP imposes a soft gradient penalty to improve upon this
(see \S\ref{relwork:stab} and \S\ref{relwork:gradpen}).
More recently, SNGAN tries to impose a 1-Lipschitz constraint by enforcing unit spectral norm of every layer.
In \S\ref{sec-comparisons}, we show that a composition of such constrained functions typically results in the overall Lipschitz constant being bounded loosely by 1, and discuss empirical results in \S\ref{sec-empirical-analysis}.

While constraining $D$ to be 1-Lipschitz is necessary for training WGANs, Miyato \etal \cite{miyato2018spectral} empirically show that such a constraint is also beneficial when training NSGANs. One may immediately see why such smoothness constraints might help. Consider the norm of $\nabla_x\mathcal{L}_G$ for NSGANs (Eq. \eqref{vanillagangrad}) and WGANs (Eq. \eqref{wgangrad}) as follows
\begin{align}
    \|\nabla_x\mathcal{L}_G \|_{\text{NSGAN}} &= \| \mathbb{E}_{x\sim \mathbb{P}_g}\left[-(1-D(x))\nabla_xf(x)\right]   \| \nonumber\\
    &\leq \mathbb{E}_{x\sim \mathbb{P}_g}\left[\|(1-D(x))\nabla_xf(x)\|\right]\nonumber \\
    &\leq \mathbb{E}_{x\sim \mathbb{P}_g}\left[\|\nabla_xf(x)\|\right]\label{eq-vanillagan-gradbound}\\
    \|\nabla_x\mathcal{L}_G \|_{\text{WGAN}} &= \|  \mathbb{E}_{x\sim \mathbb{P}_g}\left[-\nabla_xf(x)\right] \nonumber\| \\
    &\leq \mathbb{E}_{x\sim \mathbb{P}_g}\left[\|\nabla_xf(x)\|\right]\label{eq-wgan-gradbound}.
\end{align}
When $f$ is constrained to be $\mathcal{K}$-Lipschitz, clearly, $\|\nabla_x\mathcal{L}_G \|_{\text{NSGAN}}$ and $\|\nabla_x\mathcal{L}_G \|_{\text{WGAN}}$ are both bounded above by $\mathcal{K}$. This ensures that the gradients received by $G$ throughout the training are well-behaved and do not explode, thereby improving training stability.

In this paper, we introduce an alternative way to strictly bound the gradients by enforcing a piecewise $\mathcal{K}$-Lipschitz continuity almost everywhere (as opposed to a global $\mathcal{K}$-Lipschitz constraint). 
In a similar spirit to Miyato \etal \cite{miyato2018spectral}, we empirically show that our method benefits both discriminators and critics.

In the next section, we introduce a notation for the class of functions implemented by deep neural networks with piecewise linear activations, and subsequently, in \S\ref{grand}, we present gradient normalization for discriminators and critics.

\subsection{Deep piecewise linear networks}
\label{sec-pieces}
Modern deep neural networks predominantly use piecewise linear activation functions such as ReLU and LeakyReLU. Such activation functions do not admit regions with saturation, and, hence, allow training very deep networks effectively without vanishing gradients. 

Let $f(x):\mathbb{R}^{d}\rightarrow\mathbb{R}$ represent a (deterministic) deep neural network with piecewise linear activation functions. 
Denote the network parameters by $\theta$.
Then one may write
\begin{align}
    f(x,\theta) = w(x, \theta) \cdot x + b(x, \theta), \label{piecewise1}
\end{align}
where $w(x, \theta)$ and $b(x, \theta)$ are piecewise constant in $x$. We denote $w(x, \theta) \cdot x$ as the scalar dot-product of flattened tensors. 
Thus, $w(x, \theta)$ and $b(x, \theta)$ being piecewise constant in $x$ means that $\exists$ input sub-sets $S_k \in \left\{S_j\right\}_j \subseteq \mathbb{R}^{d}$, where
\begin{align}
    w(x\in S_k, \theta) &= w_k(\theta) \,~\text{and}~\, %
    b(x\in S_k, \theta) = b_k(\theta), \nonumber
\end{align}
such that $w_k$ and $b_k$ are independent of $x$ within the sub-set $S_k$. Hence, $\forall x \in S_k$, we have
\begin{align}
    f(x\in S_k, \theta) = w_k(\theta) \cdot x + b_k(\theta), \label{plane1}
\end{align}
which is linear in $x\in S_k$. 
Then $f(x,\theta)$ is the composition of continuous, piecewise linear functions, and is therefore itself a continuous and piecewise linear function of $x$. 
That is, there exist disjoint open 
input subsets $S_k$, such that $\cup_{k=1}^K S_k = \mathbb{R}^{d}$,
where Eq.\ \ref{plane1} holds.
I.e., $f(x, \theta)$ is a linear function of $x\in S_k$, with coefficients $w_k$ and $b_k$ that only depend on $\theta$.
One can interpret $w_k(\theta)$ and $b_k(\theta)$ as the \textit{effective} weights and bias of a linear functional 
(given by Eq.\  \eqref{plane1}) that equals the predictions of the deep neural network, $f(x)$,
$\forall ~ x \in S_k$. Note that, unlike a linear hyperplane in logistic regression, the effective weights $w_k(\theta)$ and bias $b_k(\theta)$ 
are only applicable for $x \in S_k$.

Given this structure, for points off of $\partial S_k$
(the boundary of $S_k$),
the gradient takes a simple form: 
\begin{align}
    \nabla_xf(x, \theta) = w_k(\theta), 
    \label{piecegradienteq}
\end{align}
which is a constant vector
$\forall ~ x \in S_k$.

\section{Gradient Normalization}
\label{grand}
In this section, we present \textit{gradient normalization} (GraN),
which strictly constrains piecewise linear networks to be 1-Lipschitz almost everywhere.

As in \S\ref{sec-pieces}, let $f(x):\mathbb{R}^{d}\rightarrow\mathbb{R}$ represent a piecewise linear neural network with parameters $\theta$. We then define the gradient normalized function $g(x)$ as  %
\begin{align}
    g(x) 
    = f(x) ~ \mathcal{R}_\epsilon(\| \nabla_xf(x)\|) = 
    \frac{f(x)~ \| \nabla_xf(x)\|}{\| \nabla_xf(x) \|^2 + \epsilon},
\end{align}
where $\epsilon>0$ is a fixed constant for numerical stability and $\mathcal{R}_\epsilon(n)=n/(n^2 + \epsilon)$
is the normalization factor, with $n=\| \nabla_xf(x)\|$.
Any bounded $R(n)$, with $R(n) = (1/n)(1 + o(1))$ as $n \rightarrow \infty$, could be tried. 
We briefly experimented with $R(n) = 1/(n + \epsilon)$, which produced similar although slightly worse results than 
$ \mathcal{R}_\epsilon $. 
The choice of $R(n)$ could benefit from further study. 
We remark that a concurrent work \cite{Wu_2021_ICCV} to ours independently explores a similar technique to regularize $D$, but with a different normalization factor.

Given this normalization, if $f$ is an arbitrary piecewise linear function then $g$ is piecewise linear such that
$ %
    \|\nabla_xg(x)\| \leq 1
$ %
analytically almost everywhere in $x\in \mathbb{R}^{d}$. While the gradient $\nabla_xg(x)$ is bounded, $g(x)$ itself can still take real values with no bounds, i.e., $g(x)\in \mathbb{R}$.

One can better describe this result with the notation developed in  \S\ref{sec-pieces}. Consider an arbitrary input $x \in S_k$ (without any loss of generality) that belongs to the input open subset $S_k$ and is mapped by the network to a linear piece given by $f(x) = w_k(\theta) \cdot x + b_k(\theta)$ $\forall ~ x \in S_k$. Then one has $\nabla_xf(x, \theta) = w_k(\theta)$ $\forall ~ x \in S_k$ (Eq. \eqref{piecegradienteq}). Therefore, the GraNed function $g(x)$, given $f(x)$, for $x\in S_k$ becomes
\begin{align}
    g(x\in S_k) =  \left[w_k(\theta) \cdot x + b_k(\theta)\right] ~\frac{\|w_k(\theta) \|}{\|w_k(\theta) \|^2 + \epsilon}.
\end{align}
Consequently, $\nabla_x g$ and its norm can be written as
\begin{align}
\nabla_x g(x\in S_k) = w_k(\theta) ~ \frac{\|w_k(\theta) \|}{\|w_k(\theta) \|^2 + \epsilon},\\
\implies \| \nabla_x g(x\in S_k)\| = \frac{\|w_k(\theta) \|^2}{\|w_k(\theta) \|^2 + \epsilon} < 1. \label{eq:nrmBnd}
\end{align}
Since $x$ and $S_k$ were arbitrarily chosen, it follows that $\| \nabla_x g(x)\| < 1$ except at the boundaries, say $x \in \partial S_k$, where the gradient does not exist.  Since $\cup_k \partial S_k$ is measure zero,
we have $\| \nabla_x g(x)\| < 1$ almost everywhere in $\mathbb{R}^{d}$.
Further, for $\|w_k(\theta) \| \gg \epsilon$, we have $||\nabla_x g|| \approx 1$ in $S_k$. 

Note that, since the original piecewise linear function $f(x)$ does not have a smooth gradient $\nabla_x f(x)$, the normalization factor $\mathcal{R}_\epsilon(\| \nabla_x f(x) \|)$ will have discontinuities for $x \in \cup_k \partial S_k$.  Therefore, $g$ is typically discontinuous and not guaranteed to be globally 1-Lipschitz.  However, we 
    empirically find that $g(x)$ has bounded finite-differences over substantial perturbations
    (see \S\ref{sec-empirical-analysis}). 
    
Due to this piecewise constant gradient property (with unit-bounded norm),
    we remark that $g$ is \textit{piecewise 1-Lipschitz}, 
    since $g$ is 1-Lipschitz with respect to any pair of points within each subset $S_k$.
It is also \textit{locally} 1-Lipschitz continuous almost everywhere,
    since there exists an open ball around every point 
    $x \in \cup_k  S_k$, 
    within which 1-Lipschitz continuity holds. Empirically, we find that the tight bound on $\| \nabla_x g(x) \|$ almost everywhere assists with GAN training. 

\subsection{Gradient Normalized GANs}
\label{granddiscsec}
Given a deep neural network represented by $f(x)$, we write $D(x)=\sigma(f(x))$ when $D$ represents discriminators, and $D(x)=f(x)$, when $D$ represents critics, respectively, where $\sigma(\cdot)$ is the sigmoid function (see \S\ref{sec-background-gans}).

When $f(x)$ is piecewise linear in $x$, we define the gradient normalized discriminator (GraND)   and critic (GraNC) as $D(x)=\sigma(g(x))$ and $D(x)=g(x)$, respectively, where 
\begin{align}
g(x) = \frac{f(x)}{\tau} ~ \frac{\| \nabla_xf(x)\|}{\| \nabla_xf(x) \|^2 + \epsilon},\label{eq:grandgx}
\end{align}
and $\tau$ is a positive constant that constrains $g$ to be $\mathcal{K}$-Lipschitz, with $\mathcal{K}=1/\tau$. For GraNDs, $\tau$ takes a role analogous to the temperature of a sigmoid, and, hence, we term it as the ``temperature" hyperparameter. 

Like spectral normalization \cite{miyato2018spectral}, the Lipschitz constant $\mathcal{K}=1/\tau$ is the only additional hyperparameter that needs to be tuned for our method. 
Moreover, in practice, our method achieves a tight bound on the 
Lipschitz constant, unlike spectral normalization, which imposes a loose upper bound with $\|g\|_{\text{Lip}}\leq\mathcal{K}$. 
See \S\ref{sec-comparisons} for details.

Finally, note that since the gradients in Eqs.\ \eqref{eq-vanillagan-gradbound} and \eqref{eq-wgan-gradbound} are computed by back-propagation, the step discontinuities in $g(x)$ are ignored.  Therefore, for GANs with GraNDs and GraNCs, 
the generator $G$ receives gradients that are always bounded, i.e., 
$\|\nabla_x \mathcal{L}_G \| \leq \mathcal{K}$.

\section{Comparison to Layer-wise Norms} %
\label{sec-comparisons}

\textbf{Spectral Normalization}~%
As noted in \S\ref{relwork:gradpen}, 
GPs softly encourage Lipschitz continuity in a data-dependent manner.
Improving on this, the spectrally normalized GAN (SNGAN) \cite{miyato2018spectral} achieves it at the architectural level.
In particular, a linear layer $\ell(x) = Wx$ (ignoring the bias term) with weights $W$ is normalized via
$ %
\widetilde{W} = { W } / { ||W||_2 },
$ %
before being applied to an input ($\widetilde{\ell}(x) = \widetilde{W}x$), 
where $ ||W||_2 = \sigma_{1}(W) $ 
is the spectral norm of $W$, equal to its largest singular value (SV) $\sigma_1$.
This is a form of WN, but acts on the whole matrix rather than its individual rows.
Notice that $|| \ell ||_\text{Lip} \leq \sigma_1(W) $ and $ \sigma_1(\widetilde{W}) = 1 $, so $ || \widetilde{\ell} ||_\text{Lip} \leq 1 $.
As such, SN ensures layer-wise 1-Lipschitz continuity, and thus enforces it across the whole network, where SVs are estimated via power iteration.\footnote{Although note that, at each step, power iteration provides a \emph{lower bound} on $\sigma_1(W)$, and thus in practice it is possible that $|| \ell ||_\text{Lip} > 1$. }

One downside of SN is the tendency to over-constrain the network, reducing capacity and attenuating gradients, due to the layer-wise enforcement mechanism \cite{anil2019sorting,li2019preventing}. 
SN guarantees a function is globally 1-Lipschitz by bounding the Lipschitz constant (LC) of every layer,
    as this then bounds their composition:
    $%
    || f \circ g ||_\text{Lip} \leq || f ||_\text{Lip} || g ||_\text{Lip}.
    $ %
However, this upper-bound is often loose, over-constraining the network and attenuating gradients (expanded upon below).
In comparison, GraN acts upon the network as a whole, leaving weights per layer free to vary, and ensuring that gradients with respect to the input are always of unit norm.

\textbf{The Looseness of Layerwise Constraints}~~%
For illustration, consider the simple case of two SNed linear layers without biases, ignoring non-linear activations (though, for instance, this occurs in the positive domain of ReLU): 
$ z = f(g(x))$, where $f(y) = By$ and $g(x) = Ax$. 
Assuming sufficient power iterations under SN, the largest singular values (SVs) of $A$ and $B$ are one, assigning  
each layer an LC of one. 
We next examine the conditions for which the composition of the layers, $f \circ g$, also has a LC of one.  

In this case, 
$z = f(g(x)) = B A x$, and so $f \circ g$ has a sharp LC of one if and only if the maximal SV of $BA$, namely $\sigma_1(BA)$, is also one.  
Let $\Gamma_\sigma(A)$ denote the span of the right singular vectors of $A$ with corresponding SVs equal to $\sigma$. We show in 
appendix \ref{appendixD},
that
$\sigma_1(BA) = 1$ if and only if the \textit{first principal angle} \cite{golub2013matrix,zhu2012angles} between the subspaces $\Gamma_1(A^T)$ and $\Gamma_1(B)$ is zero. This only occurs if they
intersect in at least one dimension.  However, if \textit{even one} SV of $A$ and $B$
is less than one, 
then $\Gamma_1(A^T)$ 
and $\Gamma_1(B)$
will be measure zero,
meaning the network must solve a high dimensional ``alignment'' problem of two measure zero sets.
The only scenario avoiding this is when every SV of either $A$ or $B$ are one, which is also a measure zero event.

Importantly, since the SN framework does not directly encourage these subspaces to align, or all the SVs of the weight matrices to be one, 
in practice it is likely that $\sigma_1 (B A) < 1$. This issue is exacerbated for deeper networks, as the overall LC equals the product of $\sigma_1 (B A)$ for every adjacent pair of layers.
In addition, though training may encourage the network to utilize its capacity by avoiding small SVs, 
empirically it struggles to do so \cite{anil2019sorting}. 
In contrast, GraN not only guarantees the function is locally 1-Lipschitz, but does so without constraining individual layers (avoiding subspace alignment issues) 
and 
enforces \textit{exactly} unit gradient almost everywhere as well 
(%
attaining the sharp LC bound within every $S_k$).
Fig.\ \ref{fig:gradnorms} displays this exactness for GraN; note that SNGAN, due its residual architecture, actually has an LC of 
1024,
showcasing the looseness of the bound.

\begin{table}[t]
  \caption{\label{tab:settings}
  Hyperparameters tested in Fig.\ \ref{fig:lrsbetascores} for GraND-GAN, SNGAN, and WGAN-GP on CIFAR-10, where $\alpha$ is the learning rate, $\beta_1$ and $\beta_2$ parametrize Adam in Eq. \eqref{eq:adam}, and $n_{\rm dis}$ is the number of discriminator steps per generator step. }
  \vspace{1mm}
  \centering
  \small{
  \begin{tabular}{lrrrr}
    \toprule
    \textbf{Setting} & $\alpha$ (LR) & $\beta_1$ & $\beta_2$ & $n_{\rm dis}$\\
    \midrule
    A & 0.0001 & 0.5 & 0.9 & 5\\
    B & 0.0002 & 0.5 & 0.999 & 1\\
    C & 0.001 & 0.5 & 0.999 & 5\\
    D & 0.001 & 0.9 & 0.999 & 5\\
    E (default) & 0.0002 & 0.0 & 0.9 & 5 \\ 
  \bottomrule
  \end{tabular}
  }
\end{table} 

\begin{figure}
	\centering
    \begin{subfigure}[b]{0.15\textwidth}
		\includegraphics[width=\textwidth]{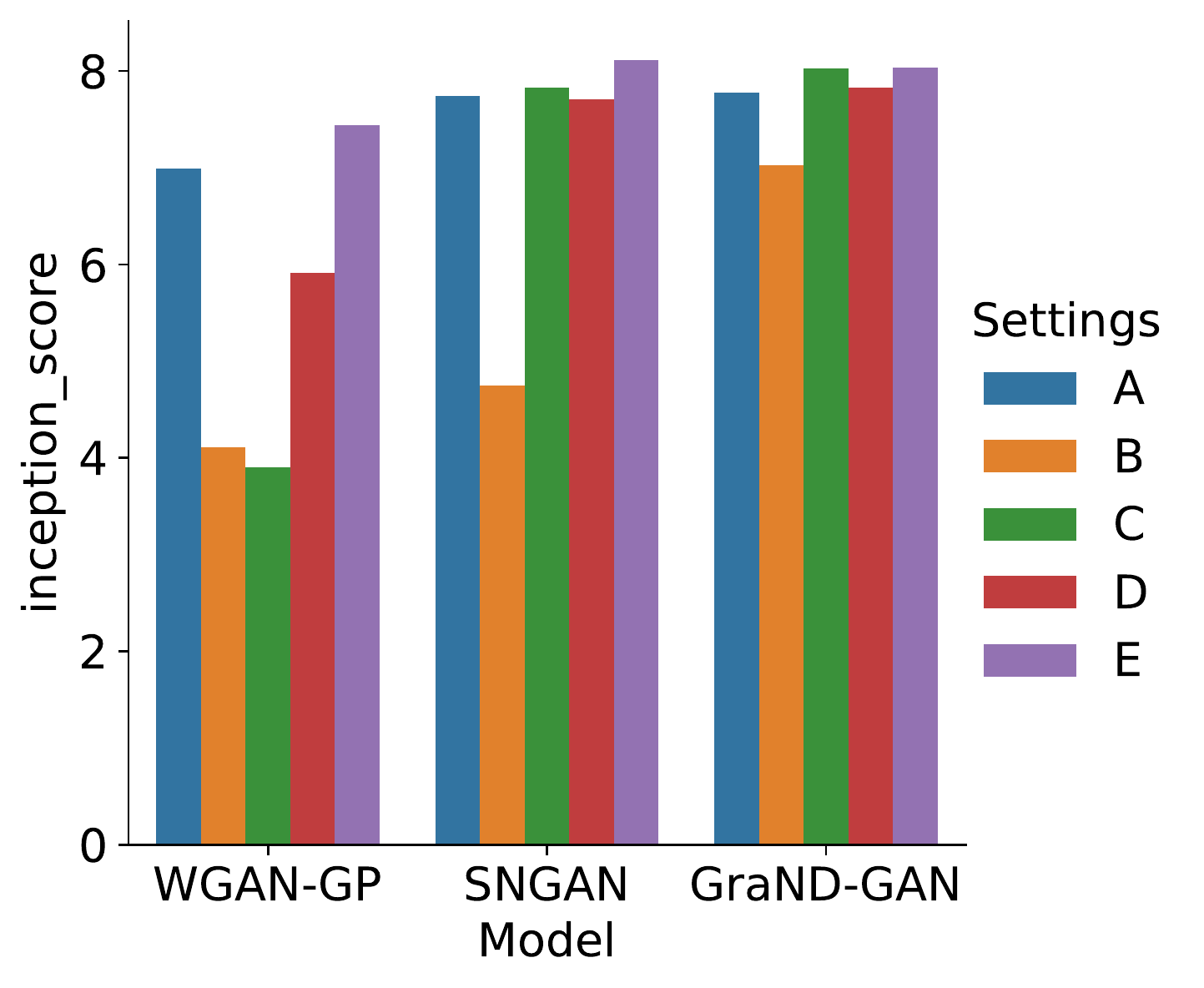}
        \caption{\label{fig:incpscores_cifar10} IS~$\uparrow$}
    \end{subfigure}%
    \begin{subfigure}[b]{0.15\textwidth}
		\includegraphics[width=1.0\textwidth]{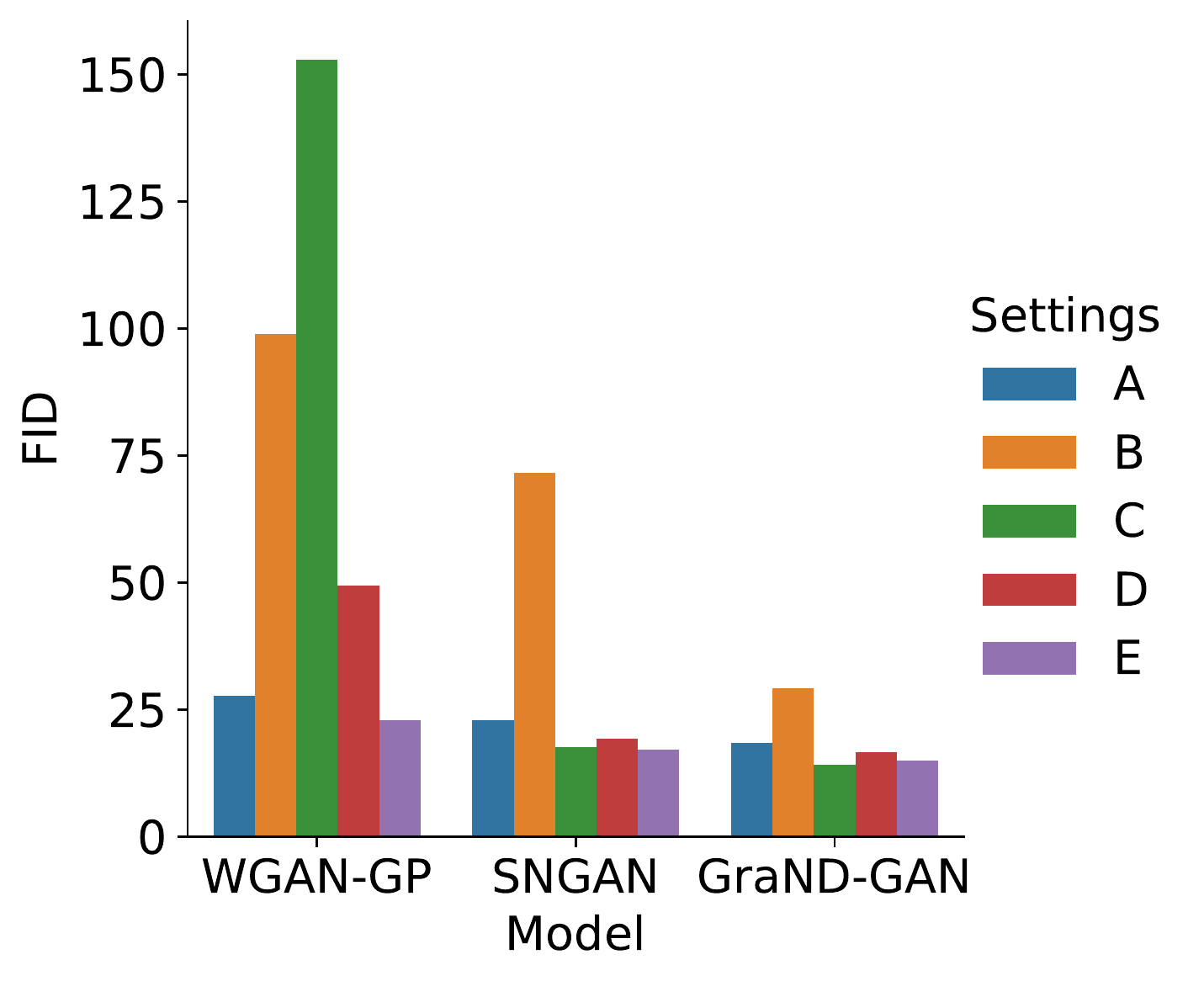}
        \caption{\label{fig:fid_cifar10} FID~$\downarrow$}
    \end{subfigure}
    \begin{subfigure}[b]{0.15\textwidth}
		\includegraphics[width=1.0\textwidth]{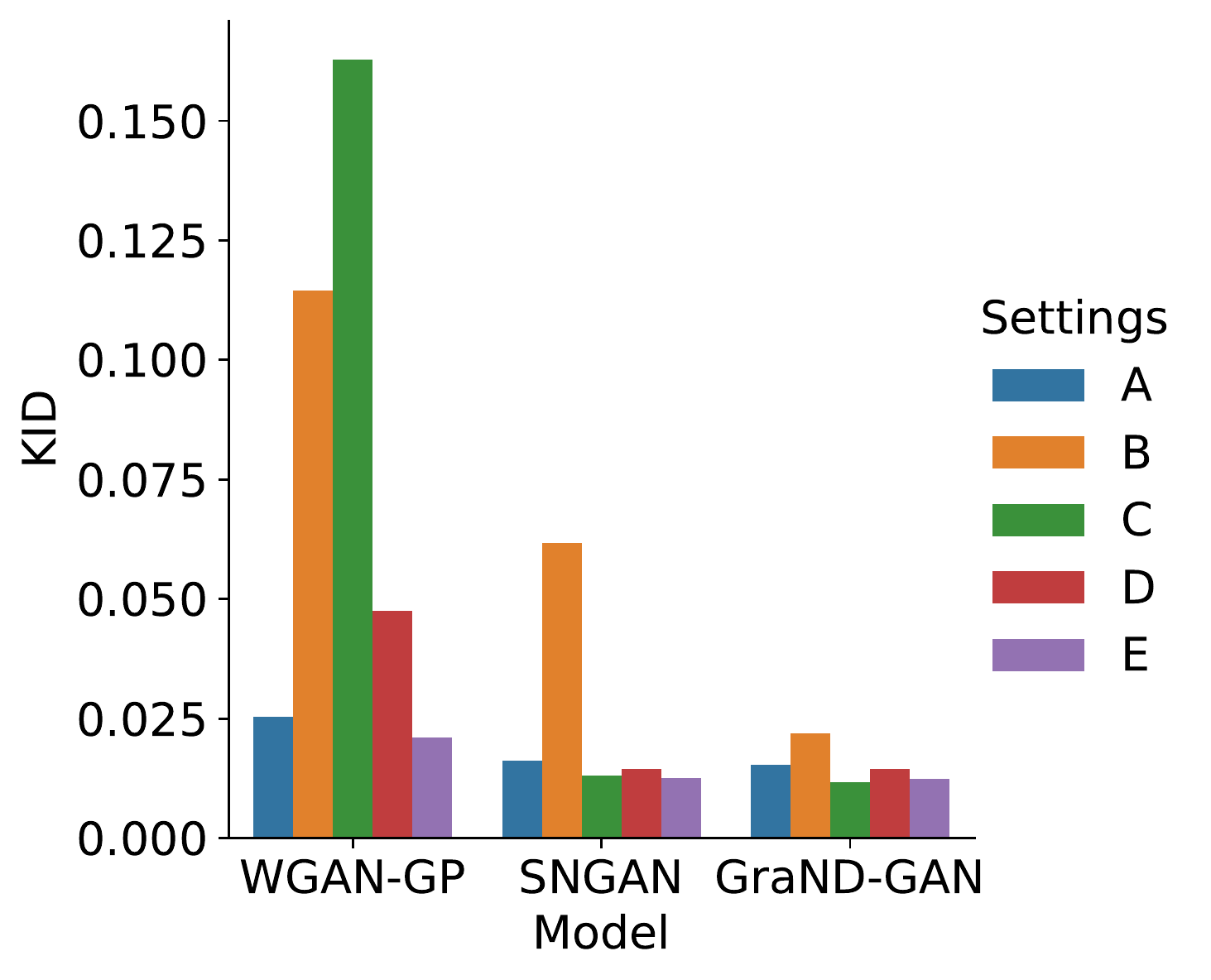}
        \caption{\label{fig:kid_cifar10} KID~$\downarrow$}
    \end{subfigure}
    \caption{\label{fig:lrsbetascores} Inception scores (IS), FIDs, and KIDs on CIFAR-10 image generation across different hyperparameters listed in Table \ref{tab:settings} for GraND-GAN, WGAN-GP, and SNGAN. Results show superior robustness for GraND. %
    }
\end{figure}

\begin{figure}
	\centering
	\includegraphics[width=0.5\textwidth]{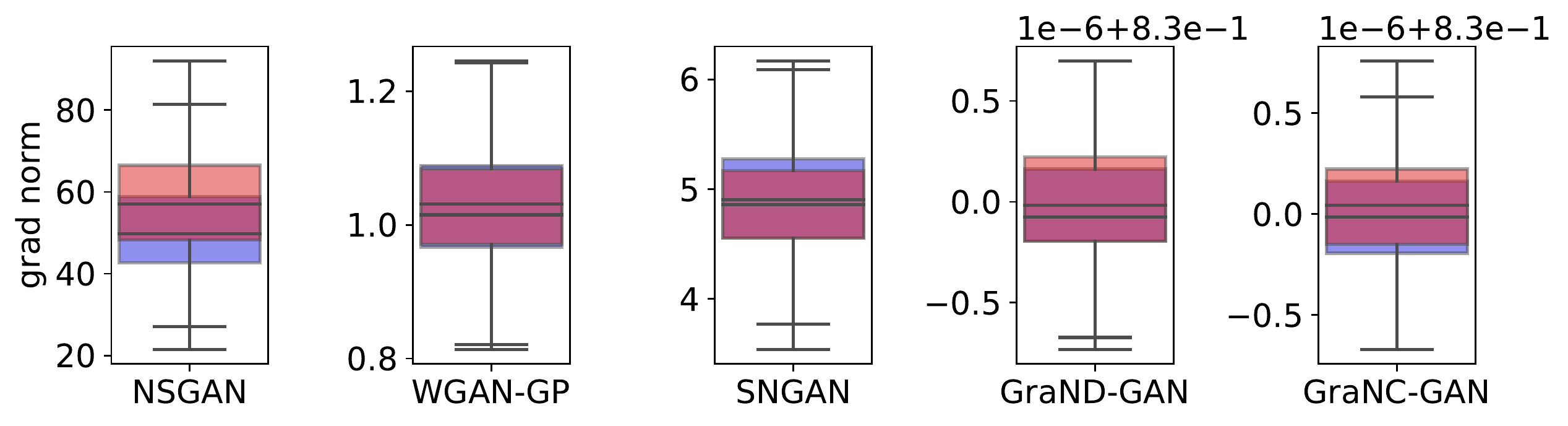}
    \caption{\label{fig:gradnorms} Boxplots of gradient norms at real (blue) and fake (red) samples for different methods at 50K iterations (out of 100K) on CIFAR-10 during training. Gradient norms for GraND/C have a very narrow distribution (spanning less than $\pm 10^{-6}$) around the piecewise Lipschitz constant $\mathcal{K}$=0.83. %
    }
\end{figure}

\section{Experiments}

\textbf{Model Architectures and Training}~%
We evaluate our method on unconditional image generation across datasets of various sizes: CIFAR-10/100 ($32\times32$) \cite{Krizhevsky09learningmultiple}, STL-10 ($48\times48$) \cite{coates2011analysis}, LSUN bedrooms ($128\times128$) \cite{yu15lsun}, and CelebA ($128\times128$) \cite{liu2015faceattributes}.  We use Mimicry \cite{lee2020mimicry} with PyTorch \cite{NEURIPS2019_9015} on a single NVIDIA V100 GPU for training our models. The generator $G$ and discriminator (or critic) $D$ architectures are identical across methods for a given dataset with identical number of learnable parameters for fair comparison. We use the Adam \cite{kingma2014adam} optimizer with $\beta_1=0.0$, $\beta_2=0.9$, and a batch size of $64$ for $100\text{K}$ iterations, with a dataset-dependent learning rate (LR) $\alpha$ and number of $D$ steps per $G$ step $n_{\rm dis}$. 
NSGANs and GraND-GAN use the cross-entropy loss for $D$ in Eq.\  \eqref{eqD1} and the non-saturating loss for $G$ in Eq. \eqref{eqG2}. 
GraNC-GAN uses a soft version of the hinge loss for $D$ in Eq. \eqref{wgand2}, replacing ReLU with softplus, and Eq.\ \eqref{wgang1} for $G$ (see \S\ref{sec-background-gans}). %
The soft hinge loss was found to be more performant and stable than the (hard) hinge loss with GraNC-GANs: on LSUN, GraNC diverged with the hard hinge loss, while on CelebA, a lower FID was obtained. 
See appendix for further model details, ablation experiments, and hyperparameter choices. \looseness=-1

\textbf{Lipschitz Constant Analysis}~~%
We find that tuning the Lipschitz constant $\mathcal{K}$ significantly affects the performance and stability of models when using the Adam optimizer. 
This is due to the interdependence of $\epsilon_{\text{Adam}}$ in the update 
with $\mathcal{K}$:
\begin{equation}  \delta \mathbf{\theta} = - \frac{\langle \mathbf{g} \rangle_{\beta_1}}{\sqrt{\langle\mathbf{g}^2\rangle_{\beta_2}} + \epsilon_{\text{Adam}}}. \label{eq:adam} \end{equation}
Changing $\mathcal{K}$ from its default value of one has an effect of scaling the gradients of the loss function $\mathbf{g}$ by $\mathcal{K}$. This in turn has an effect of scaling    $\epsilon_{\text{Adam}} \longrightarrow \epsilon_{\text{Adam}}/\mathcal{K}$. Empirically, we find that the Adam update for individual parameters can go $\ll 10^{-7}$ in magnitude on the plateaus of the loss landscape where $\epsilon_{\text{Adam}}$ becomes significant. We find it helps having smaller $\mathcal{K}$ generally when training on larger image resolutions which has an effect of increasing $\epsilon_{\text{Adam}}$ from its default value of $1\times 10^{-8}$. Intuitively, a larger  $\epsilon_{\text{Adam}}$ suppresses the Adam update when the loss gradient magnitudes are $\lesssim \epsilon_{\text{Adam}}$. Therefore, $\epsilon_{\text{Adam}}$ determines the extent of noisy Adam updates at plateaus of the loss landscape, and so may need tuning. More details on the choice of $\mathcal{K}$ are provided in the appendix. \looseness=-1

\textbf{Baselines and Evaluation}~~%
To directly compare our normalization method GraN with gradient penalty (GP) and spectral normalization (SN), independent of the loss function, we train NSGAN with a gradient penalty $P_1$ loss (NSGAN-GP), and with SNed layers (NSGAN-SN). We also train NSGAN-GP$\dagger$ and NSGAN-SN$\dagger$, which correspond to models constrained with a tuned Lipschitz constant (instead of 1). WGAN-GP$\dagger$ and SNGAN$\dagger$ are defined analogously.

We quantitatively evaluate the methods by Inception Score (IS) \cite{salimans2016improved}, FID \cite{fid}, and KID \cite{binkowski2018demystifying} with 50K synthetic images randomly sampled from $G$ and 50K real images from the dataset. 
IS is not used for LSUN and CelebA, as these comprise a single class, for which IS performs poorly \cite{lee2020mimicry}. See appendix \ref{appendixA} for additional details.

\textbf{Unconditional Image Generation}~~%
Table \ref{tab:cifar10-100-results} presents the comparative results on CIFAR-10, CIFAR-100, and STL-10 image generation. Our methods rank among the top two across every metric (KID, FID, IS) on CIFAR-10 and CIFAR-100. On STL-10, our two methods rank the highest in IS and FID, however, we fall behind on KID by a small margin compared to SNGAN and SNGAN$\dagger$.\looseness=-1

Table \ref{tab:lsunceleba-results} summarizes our results on LSUN bedrooms and CelebA image generation. 
The NSGAN model did not converge (i.e., FID $> 70$) in two random restarts of training for both LSUN and CelebA. 
Similarly, WGAN-GP failed to converge on CelebA in two runs. 
GraND-GAN achieves the best results on CelebA, and the second-best on LSUN bedrooms, falling slightly behind NSGAN-GP$\dagger$. 
Among critics, GraNC-GAN performs best by FID as well.

\begin{table*}[t]
\centering
  \caption{\label{tab:cifar10-100-results}Inception scores (IS), FIDs, and KIDs with unsupervised image generation on CIFAR-10, CIFAR-100, and STL-10.
  The best and the second best models per evaluation metric and GAN family (i.e., with discriminators or critics) are indicated by \boldred{bold red} and \boldblue{bold blue} fonts. 
  $\dagger$ indicates modified baselines with an altered Lipschitz constant $\mathcal{K}$. The table is split comparing discriminators (top) and critics (bottom). We write ``\textbf{--}" for cases where a model did not achieve a FID $< 70$.
  }\vspace{-2mm}
  \resizebox{\textwidth}{!}{
  \begin{tabular}[t]{lrrrrrrrrrr}
    \toprule
    \multirow{2}{*}{Method} & \multicolumn{3}{c}{\textbf{IS}~$\uparrow$} & \multicolumn{3}{c}{\textbf{FID}~$\downarrow$} & \multicolumn{3}{c}{\textbf{KID} (${\times}1000$) ~$\downarrow$} \\
    & CIFAR-10 & CIFAR-100 & STL-10 & CIFAR-10 & CIFAR-100 & STL-10 & CIFAR-10 & CIFAR-100 & STL-10\\ 
    \midrule
    NSGAN & {7.655} & {6.611} & {7.920} & {23.750} & {30.842} & {44.179} & {14.5} & {20.5} & {40.0}  \\
    NSGAN-GP & {8.016} & \textbf{--} & {8.568} & \boldblue{15.813} & \textbf{--} & {38.848} & \boldblue{12.9} & \textbf{--} & {38.9}  \\
    NSGAN-SN & {7.792} & {7.258} & {8.167} & {20.998} & {25.564} & \boldblue{38.669} & {15.7} & {18.4} & \boldblue{35.7}  \\
    NSGAN-GP$\dagger$ & \boldblue{8.019} & \boldblue{7.892} & \boldblue{8.623} & {15.911} & \boldblue{20.894} & {40.110} & {13.1} & \boldblue{17.0} & {39.8} \\
    NSGAN-SN$\dagger$ & {7.814} & {7.526} & {8.135} & {20.323} & {24.200} & {39.013} & {15.3} & {17.7} & {36.7} \\
    GraND-GAN (Ours)  & \boldred{8.031} & \boldred{8.314} & \boldred{8.743} & \boldred{14.965} & \boldred{18.978} & \boldred{35.226} & \boldred{12.3} & \boldred{13.7} & \boldred{35.0}  \\
    \midrule \midrule
    WGAN-GP & {7.442} & {7.520} & {8.492} & {22.927} & {27.231} & {42.170} & {21.1} & {23.5} & {43.0} \\
    SNGAN & \boldred{8.112} & {7.778} & {8.385} & {17.107} & {20.739} & {38.218} & \boldblue{12.6} & \boldred{14.3} & \boldblue{34.3} \\
    WGAN-GP$\dagger$ & {7.344} & {7.684} & {8.466} & {22.705} & {25.211} & {42.595} & {20.6} & {21.2} & {44.7} \\
    SNGAN$\dagger$ & \boldblue{7.991} & \boldblue{7.959} & \boldblue{8.552} & \boldblue{16.740} & \boldblue{20.104} & \boldblue{36.203} & \boldred{12.0} & \boldred{14.3} & \boldred{33.3} \\
    GraNC-GAN (Ours)  & {7.966} & \boldred{8.208} & \boldred{8.957} & \boldred{16.361} & \boldred{19.131} & \boldred{35.770} & {13.7} & \boldblue{14.8} & {35.4} \\
    \bottomrule
  \end{tabular}
  }
\end{table*}

\begin{table}[t]
\centering
  \caption{\label{tab:lsunceleba-results} Unsupervised $128\times128$ image generation on LSUN-Bedrooms and CelebA. We write ``\textbf{--}" to indicate the cases where a model did not achieve a FID $< 70$ in two random training restarts. The best and the second best models per evaluation metric and GAN family (i.e., with discriminators or critics) are indicated by \boldred{bold red} and \boldblue{bold blue} fonts. 
  $\dagger$ indicates modified baselines with an altered Lipschitz constant $\mathcal{K}$.
  We split the table into discriminators (top) and critics (bottom), to better highlight the differences per loss function. 
  GraN performs best or second-best across all datasets, losses, and performance metrics; in the case of discriminators, GraND-GAN and NSGAN-GP$\dagger$ (our Lipschitz-tuned GP-based approach) are the top two performers.
  }\vspace{1mm}
  \resizebox{0.47\textwidth}{!}{
  \begin{tabular}[]{lrrrrrrr}
    \toprule
    \multirow{2}{*}{Method} & \multicolumn{2}{c}{\textbf{FID}~$\downarrow$} & \multicolumn{2}{c}{\textbf{KID (${\times}1000)$}~$\downarrow$} \\
    & LSUN & CelebA & LSUN & CelebA\\
    \midrule
    NSGAN  & \textbf{--} & \textbf{--} & \textbf{--} & \textbf{--}  \\ %
    NSGAN-GP  & \textbf{--} & \textbf{--} & \textbf{--} & \textbf{--}  \\
    NSGAN-SN  & {74.926} & {14.33} & {44.8} & {21.2}  \\
    NSGAN-GP$\dagger$  & \boldred{10.483} & \boldblue{9.385} & \boldred{7.2} & \boldblue{5.8} \\
    NSGAN-SN$\dagger$  & {12.635} & {9.644} & {8.3} & \boldblue{5.8} \\
    GraND-GAN (Ours)   & \boldblue{10.795} & \boldred{9.377} & \boldblue{7.3} & \boldred{5.2}  \\
    \midrule \midrule
    WGAN-GP  & {13.562} & \textbf{--} & {9.8} & \textbf{--} \\ %
    SNGAN  & \boldblue{13.237} & \boldblue{13.466} & \boldred{8.0} & {8.9} \\
    WGAN-GP$\dagger$  & {16.884} & \textbf{--} & {12.0} & \textbf{--} \\
    SNGAN$\dagger$  & {67.346} & {15.874} & {32.0} & \boldblue{8.7}\\
    GraNC-GAN (Ours)  & \boldred{12.533} & \boldred{12.000} & \boldblue{9.1} & \boldred{8.1} \\ 
    \bottomrule
  \end{tabular}
  }
\end{table}

Figure \ref{fig:lrsbetascores} presents a comparison of GraND-GAN with WGAN-GP and SNGAN on CIFAR-10 image generation across various training settings listed in Table \ref{tab:settings}. For setting B ($n_{\rm dis}$ = 1), our method retains a respectable FID score (and other metrics) compared to SNGAN and WGAN-GP. For settings C and D with larger learning rates and momentum hyperparameters, the performance of WGAN-GP degrades while our method and SNGAN remain quite robust.

\begin{figure}
	\centering
    \begin{subfigure}[b]{0.23\textwidth}
		\includegraphics[width=1.0\textwidth]{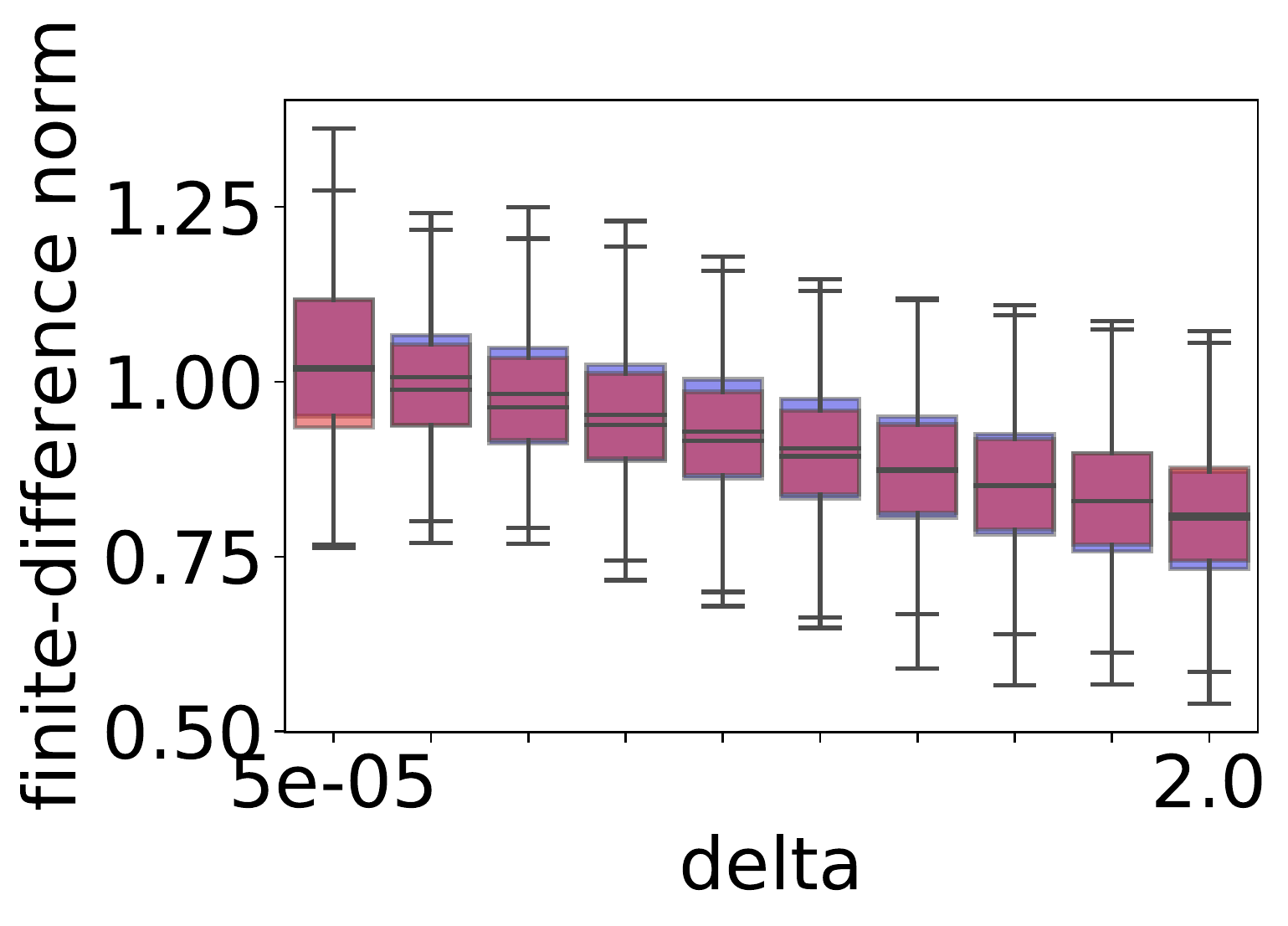}
        \caption{\label{fig:fid_cifar10:w} WGAN-GP}
    \end{subfigure}
    \begin{subfigure}[b]{0.215\textwidth}
		\includegraphics[width=1.0\textwidth]{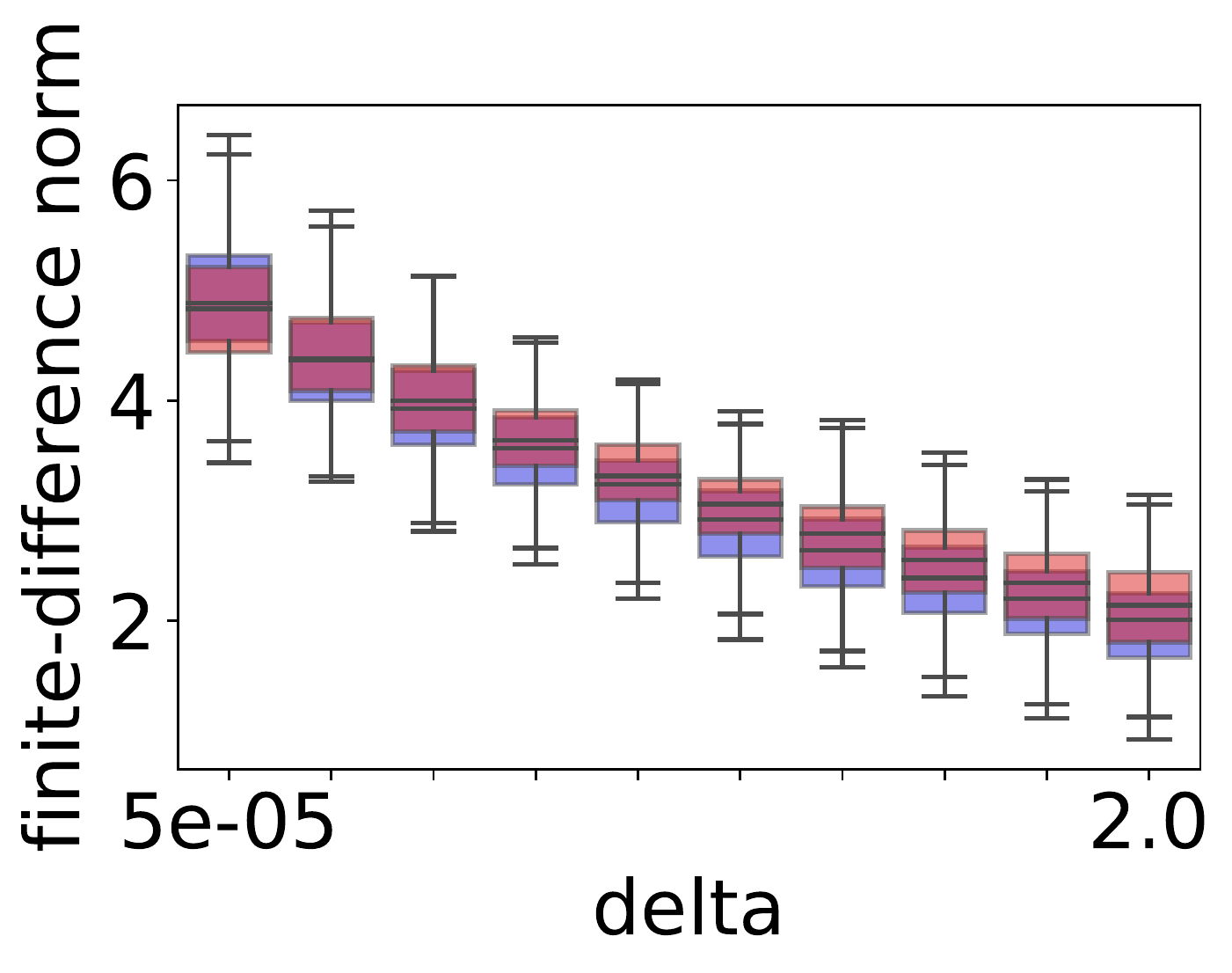}
        \caption{\label{fig:fid_cifar10:s} SNGAN}
    \end{subfigure}
    \begin{subfigure}[b]{0.22\textwidth}
		\includegraphics[width=1.0\textwidth]{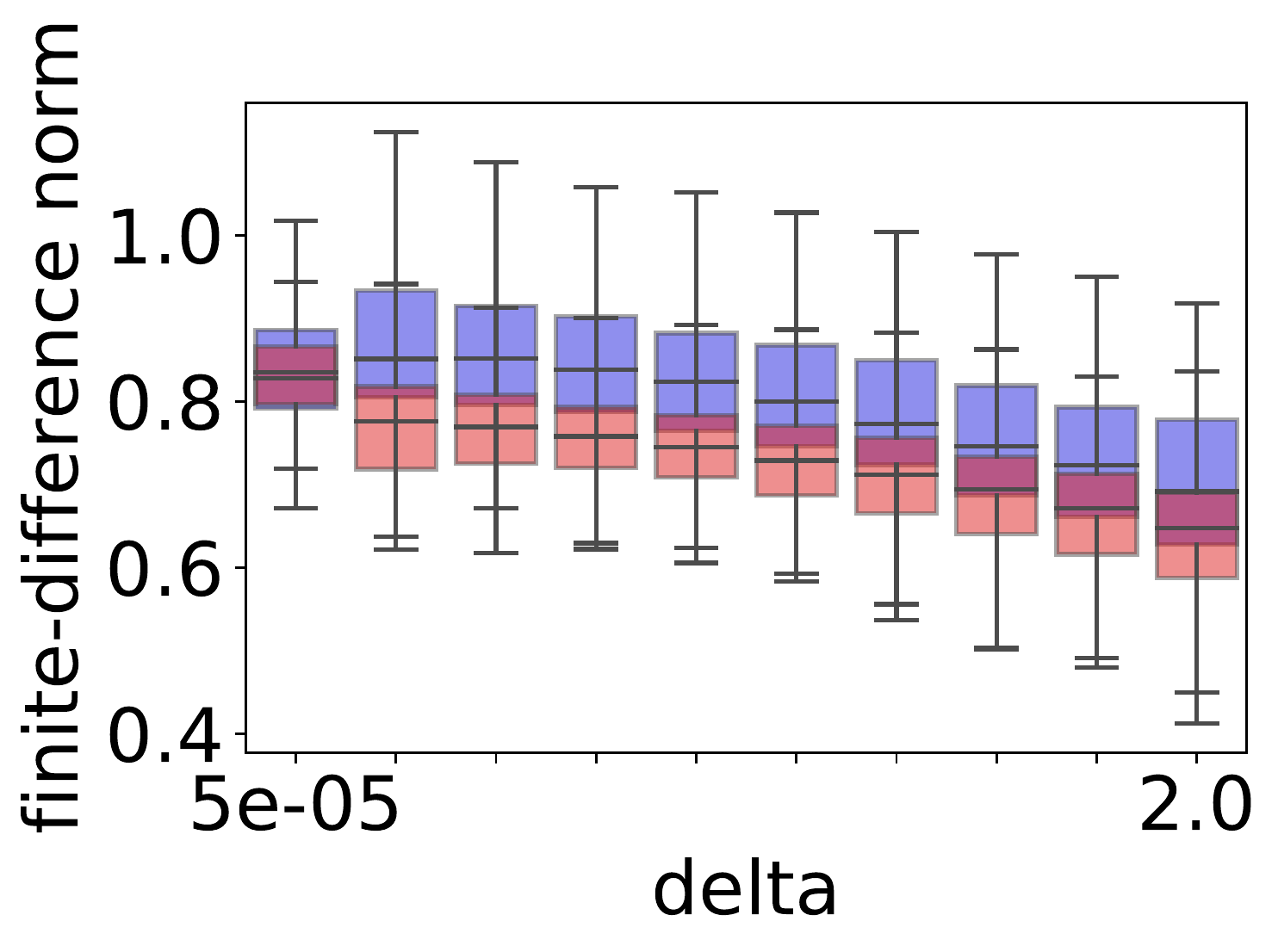}
        \caption{\label{fig:fid_cifar10:gd} GraND-GAN}
    \end{subfigure}
    \begin{subfigure}[b]{0.22\textwidth}
		\includegraphics[width=1.0\textwidth]{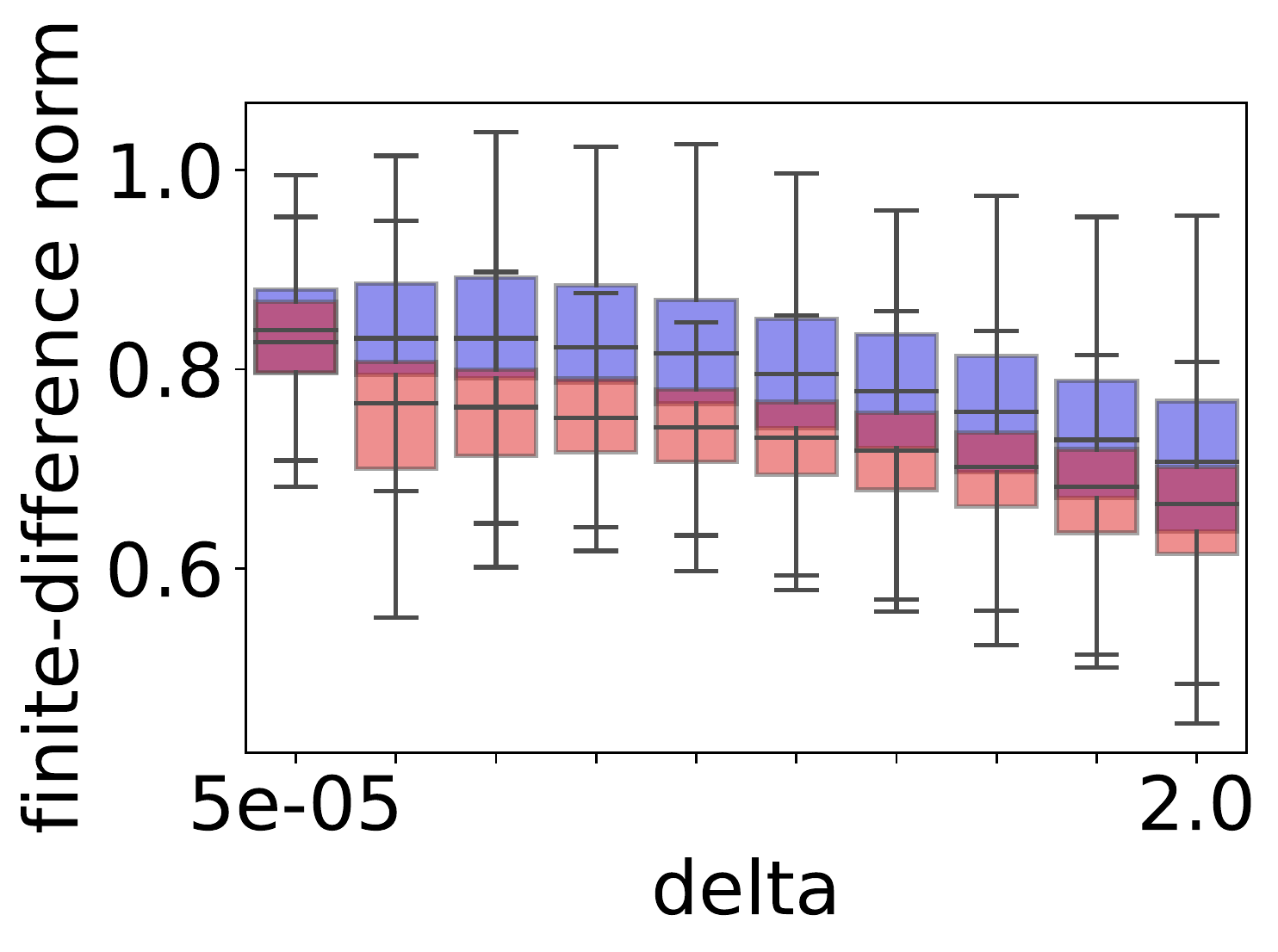}
        \caption{\label{fig:fid_cifar10:gc} GraNC-GAN}
    \end{subfigure}
    \caption{\label{fig:fdgradnorms} Boxplots of estimated finite-difference gradient norm with increasing $L2$ perturbation strengths $\delta$ around the real (blue) and fake samples (red) on CIFAR-10 at 50K iterations (out of 100K) for different methods. Empirically, GraND/C is fairly Lipschitz bounded across polytopes globally, close to the piecewise Lipschitz constant of $\mathcal{K}=0.83$. 
    }
\end{figure}

\section{Gradient Normalization Empirical Analysis}

\label{sec-empirical-analysis}

In this section, we empirically analyze the effect of gradient normalization on (a) the gradient norms and (b) the finite-difference approximation to the gradient norm at increasing levels of perturbation $\delta$, and compare it with spectral normalization and gradient penalty. 

Figure \ref{fig:gradnorms} shows a boxplot of $\|\nabla_xf(x)\|$  for the baselines, and $\|\nabla_xg(x)\|$ for our methods, on a CIFAR-10 image generation task at 50K iterations (out of 100K), where $f(x)$ is the piecewise linear discriminator (or critic) network and $g(x)$ is its gradient normalized version. The gradient norms $\| \nabla_xf(x)\|$ for NSGAN with an unconstrained discriminator are substantially larger.   For WGAN-GP and SNGAN,  $\|\nabla_xf(x)\|$ are bounded within a reasonable range. Gradient normalized discriminators and critics with a piecewise Lipschitz constant of $\mathcal{K}=0.83$, have a narrow distribution with a gradient norm that is $\approx\mathcal{K}$ ($\pm 10^{-6}$) across samples. %

Unlike spectral normalization or gradient penalty, our method does not constrain the discriminator or critic to be globally $\mathcal{K}$-Lipschitz. We investigate this on a CIFAR-10 image generation task by first sampling fake data from $G$ at a given training iteration and real data from the dataset. We estimate $\Delta_i = \| h(x_i + \delta n_i) - h(x_i)\|/\delta$ for each of the samples, which is the finite difference along a step of magnitude $\delta>0$ along the local gradient direction $n_i = \nabla h(x_i)/ \| \nabla h(x_i) \|$, where $h$ denotes $f$ for baselines and $g$ for our methods, respectively. This provides a probe for the LC (albeit a lower bound) similar in spirit to prior work \cite{zou2019lipschitz}.

Figure \ref{fig:fdgradnorms} shows a boxplot of the resulting $\Delta_i$ for increasing perturbation magnitudes $\delta$. Evidently, gradient normalized discriminators and critics have  well-behaved finite-differences, even for fairly large neighborhoods $\delta$, despite being only piecewise Lipschitz in theory. Moreover, the variance of the computed $\Delta_i$'s across $\delta$ for our methods is comparable to WGAN-GP.

\section{Discussion}

\textbf{Limitations}~~We observed instabilities when training gradient normalized critics with the Wasserstein loss (Eq.\ \eqref{wgand1}). 
The hinge loss (Eq.\ \eqref{wgand2}) improved this, but still struggled for larger images; the soft hinge approach was found to work better.
However, training GraND with the NS loss (Eq.\ \eqref{eqD1}) was found to be more stable, especially on larger images.
Also, on such images, our method periodically diverged late in training, an issue present for the baselines as well.

\textbf{Future work}~~While GraN does not guarantee a global Lipschitz constraint due to discontinuities, it does enforce constant bounded-norm gradients almost everywhere (and thus piecewise Lipschitz continuity). Moreover, empirically, it is competitive with, or better than, existing baselines. 
Investigating global versus local Lipschitz continuity, as well as gradient regularization,
is thus an enticing future direction.

\textbf{Conclusion}~~We introduced a novel input-dependent normalization for piecewise linear critics and discriminators. Our method guarantees a bounded input gradient norm almost everywhere and is piecewise $\mathcal{K}$-Lipschitz. We empirically showed that our method improves unconditional image generation using GANs across a range of datasets. Finally, though our method does not explicitly impose a global $\mathcal{K}$-Lipschitz constraint, empirically, the finite-difference gradient norm is well-behaved in a large local neighbourhood. %

{\small
\bibliographystyle{ieee_fullname}
\bibliography{paper}
}

\clearpage

\section*{Appendix}

\appendix
\section{Training Details}
\label{appendixA}

We use Mimicry \cite{lee2020mimicry} with PyTorch \cite{NEURIPS2019_9015} on a single NVIDIA V100 GPU for training our models. The generator $G$ and discriminator (or critic) $D$ architectures are identical across methods for a given dataset except for models with spectral normalization that replace convolutional and linear layers with their normalized variants. The number of learnable parameters are identical across methods for a fixed dataset size. Number of parameters for $(G, D)$ are $\approx$ $(4.3\text{M}, 1\text{M})$ for $32^2$, $(4.9\text{M}, 10\text{M})$ for $48^2$, and $(32\text{M}, 29\text{M})$ for $128^2$ image sizes, respectively. $G$ and $D$ are both residual networks with ReLU activation functions, and $G$ employs batch normalization \cite{ioffe2015batch} while $D$ does not. We train our models on a single NVIDIA V100 GPU with the Adam \cite{kingma2014adam} optimizer at a learning rate (LR) of $2\times10^{-4}$, $\beta_1=0.0$, $\beta_2=0.9$ and a batch size of $64$ for $100\text{K}$ iterations. The number of discriminator updates per generator update $n_{dis}$ is set to $5$ for CIFAR-10/CIFAR-100/STL-10 and $2$ for LSUN bedrooms/CelebA. All models (GraN or baseline) use a linear LR decay policy except models on CelebA that use the same learning rate throughout, following Mimicry \cite{lee2020mimicry}. However, GraNC-GAN on CelebA required a slight alteration: setting LRs for $G$ and $D$ to be $5\times10^{-5}$ and $1\times10^{-4}$, respectively, and using linear LR decay. %

Empirically we find it necessary to have a smaller piecewise Lipschitz constant $\mathcal{K}$ when training GANs on larger image resolutions with gradient normalization. We suspect that a smooth discriminator or critic with smaller gradient norms is essential for stable GAN training on larger image resolutions. We choose $\mathcal{K}=1/\tau=0.0909$ for our models on LSUN bedrooms/CelebA (except for $\mathcal{K} = 1/\tau = 0.2$ with GraNC-GAN on CelebA) and $\mathcal{K}=1/\tau=0.83$ for our models on CIFAR-10/CIFAR-100/STL-10.

WGAN-GP uses the Wasserstein distance based loss objectives for $D$ in Eq. (7) and $G$ in Eq. (8). SNGAN uses hinge loss for $D$ in Eq. (10) and Eq. (8) for $G$.

For NSGAN-GP$\dagger$, we adjust the gradient penalty loss to constrain the Lipschitz constant to $\mathcal{K}$ (instead of 1). For NSGAN-SN$\dagger$, we scale the output of the network before the sigmoid by $\mathcal{K}$ to obtain an effective $\mathcal{K}$-Lipschitz constraint using SN. We also retrain the baselines WGAN-GP and SNGAN with similar modifications so that the Lipschitz constraint is identical to the piecewise Lipschitz constraint for our methods and call them WGAN-GP$\dagger$ and SNGAN$\dagger$, respectively. 

It is also worth highlighting that our method backpropagates through the GraN normalization term as well and does not simply treat it as a constant.

\paragraph{Evaluation}~%
We quantitatively evaluate the methods by Inception Score (IS) \cite{salimans2016improved}, FID \cite{fid}, and KID \cite{binkowski2018demystifying} with 50K synthetic images randomly sampled from $G$ and 50K real images from the dataset. 
We report the mean scores computed across 3 randomly sampled sets of 50K images for a given $G$. We note that across all methods and datasets, the standard deviations across 3 evaluation samplings for IS, FID, and KID are less than 0.05, 0.085, and 0.0004, respectively, and we therefore do not include them in our tables. 
IS is not used for LSUN and CelebA, as these comprise a single class, for which IS performs poorly \cite{lee2020mimicry}.

\section{Model Architectures}
\label{appendixB}

\begin{figure}
	\centering
    \begin{subfigure}[b]{0.105\textwidth}
		\includegraphics[width=1.0\textwidth]{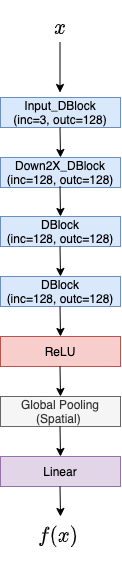}
        \caption{$x\in\mathbb{R}^{32^2}$}
    \end{subfigure} \hfill
    \begin{subfigure}[b]{0.105\textwidth}
		\includegraphics[width=1.0\textwidth]{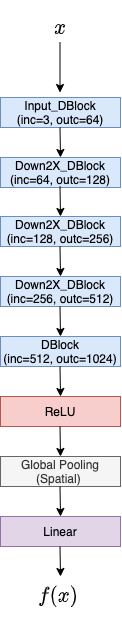}
        \caption{$x\in\mathbb{R}^{48^2}$}
    \end{subfigure}\hfill
    \begin{subfigure}[b]{0.105\textwidth}
		\includegraphics[width=1.0\textwidth]{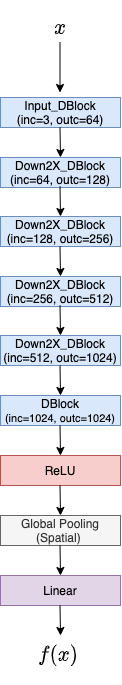}
        \caption{$x\in\mathbb{R}^{128^2}$}
    \end{subfigure}
    \caption{\label{fig:discarch} Discriminator architectures for a) $32\times32$, b) $48\times48$ and c) $128\times128$ image sizes, respectively. The architectures for Input\_DBlock, DBlock and Down2X\_DBlock are described in Figure \ref{fig:blockarchs}. All models use Global Spatial Average Pooling except SNGAN that uses Global Spatial Sum Pooling before the last Linear layer. For SNGAN only, the \texttt{Linear} and convolution \texttt{Conv2D} layers are the spectral normalized versions with 1 power iteration.    %
    }
\end{figure}

\begin{figure}
	\centering
    \begin{subfigure}[b]{0.15\textwidth}
		\includegraphics[width=1.0\textwidth]{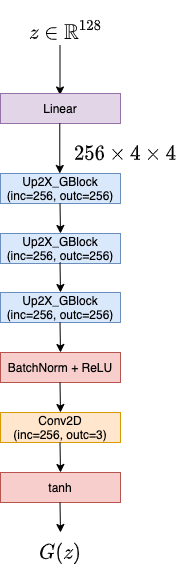}
        \caption{$G(z)\in\mathbb{R}^{32^2}$}
    \end{subfigure} \hfill
    \begin{subfigure}[b]{0.15\textwidth}
		\includegraphics[width=1.0\textwidth]{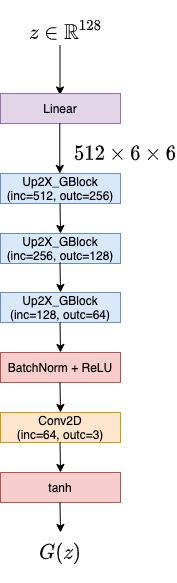}
        \caption{$G(z)\in\mathbb{R}^{48^2}$}
    \end{subfigure}\hfill
    \begin{subfigure}[b]{0.15\textwidth}
		\includegraphics[width=1.0\textwidth]{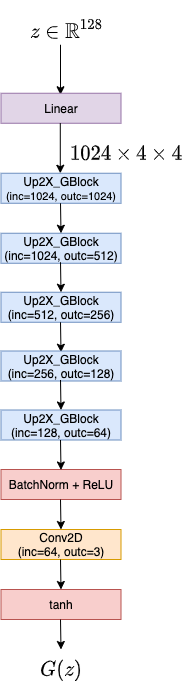}
        \caption{$G(z)\in\mathbb{R}^{128^2}$}
    \end{subfigure}
    \caption{\label{fig:genarch} Generator architectures for a) $32\times32$, b) $48\times48$ and c) $128\times128$ image sizes, respectively. The architecture for Up2X\_GBlock is described in Figure \ref{fig:blockarchs}. Generator architectures are identical across all models for a given dataset resolution. %
    }
\end{figure}

Figure \ref{fig:discarch} presents the discriminator model architectures for inputs of dimensions $32^2$, $48^2$ and $128^2$, respectively. Figure \ref{fig:genarch} presents the generator model architectures for outputs of dimensions $32^2$, $48^2$ and $128^2$, respectively.

Note that for GraN-models, the output of the networks $f(x)$ is normalized to $g(x)$ as described in Equation (19) of the main paper and does not contain any additional learnable parameters.

To modify the Lipschitz constant (LC) of baselines involving spectral normalized linear and convolutional layers (NSGAN-SN$\dagger$ and SNGAN$\dagger$), we scale the output $f(x)$ by $\mathcal{K}$, i.e., $f(x)\rightarrow\mathcal{K}f(x)$, where $\mathcal{K}$ scales LC relative to the LC of the baseline model since $|f|_{\text{Lip}}\leq1$, implies, $|\mathcal{K}f|_{\text{Lip}}\leq\mathcal{K}$.

For models WGAN-GP$\dagger$ and NSGAN-GP$\dagger$, we instead only change the gradient penalty loss term in the objective for $D$ to
\[\mathcal{L}_{\text{GP}} = \lambda\left(\| \nabla_x f(x) \| - \mathcal{K} \right)^2, \]
where $\lambda=10$ (following defaults recommended in \cite{gulrajani2017improved}) and $\mathcal{K}=1$ corresponds to the default WGAN-GP model. As in the main paper, we denote $\dagger$ to represent models with adjusted LC relative to the original baselines. 

See also \S\hyperref[appendixD]{D} and Fig.\ \ref{fig:gradnormsappendix} for discussion and empirical results concerning the observed LC when using an SNGAN discriminator with a resnet-based convolutional architecture.

\begin{figure}
	\centering
    \begin{subfigure}[b]{0.22\textwidth}
		\includegraphics[width=1.0\textwidth]{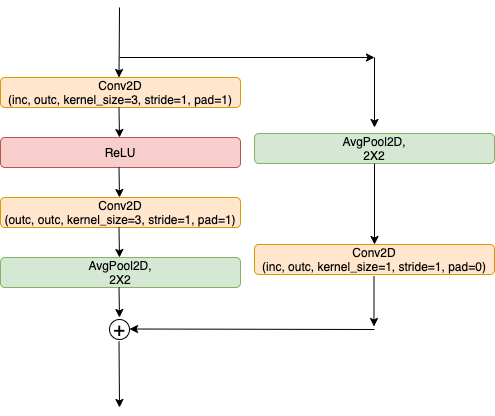}
        \caption{Input\_DBlock}
    \end{subfigure} \hfill
    \begin{subfigure}[b]{0.22\textwidth}
		\includegraphics[width=1.0\textwidth]{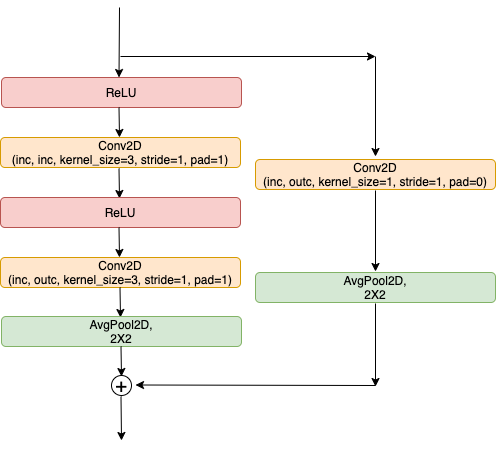}
        \caption{Down2X\_DBlock}
    \end{subfigure}\\
    \begin{subfigure}[b]{0.15\textwidth}
		\includegraphics[width=1.0\textwidth]{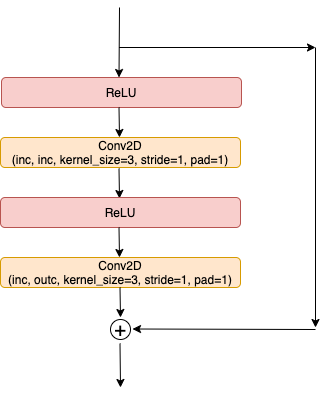}
        \caption{DBlock}
    \end{subfigure}\hfill
    \begin{subfigure}[b]{0.22\textwidth}
		\includegraphics[width=1.0\textwidth]{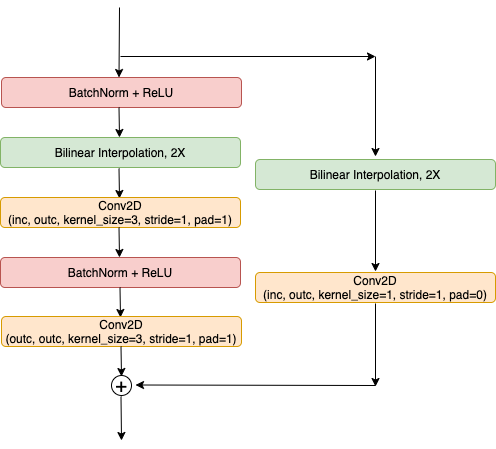}
        \caption{Up2X\_GBlock}
    \end{subfigure}
    \caption{\label{fig:blockarchs} Residual block architectures for a) Input\_DBlock, b) Down2X\_DBlock, c) DBlock  and d) Up2X\_GBlock in Figures \ref{fig:discarch} and \ref{fig:genarch}. \texttt{inc} and \texttt{outc} denote the input and output number of channels, respectively. Note that when $\texttt{inc}\neq\texttt{outc}$, the skip connection in DBlock includes a $1\times 1 ~\texttt{Conv2D}$ appropriately. For SNGAN, the linear and convolution layers in Input\_DBlock, DBlock and Down2X\_DBlock are the spectral normalized versions.}
\end{figure}

\section{Wall-clock timings for a single training update} \label{appendixC}

We summarize wall-clock times for a single training update that consists of one generator update and $n_{dis}$ number of discriminator updates (including time for loading a batch of $64$ images from the dataset). As can be noticed from Table \ref{tab:wallclock}, our method is roughly similar in wall-clock timings compared to WGAN-GP on smaller models ($32^2$ and $48^2$) but slower than NSGAN or SNGAN. On $128^2$ images GraN is 40\% slower than WGAN-GP. 

\begin{table}[h]
\centering
  \caption{\label{tab:wallclock} Wall-clock timings (in seconds $\times 10$) for a single training update across different dataset of different resolutions. Note that $n_{dis}=5$ for CIFAR-10/100 and STL-10 while $n_{dis}=2$ for LSUN/CelebA, following Mimicry \cite{lee2020mimicry}.
  }\vspace{1mm}

 \resizebox{0.47\textwidth}{!}{
  \begin{tabular}[]{lrrrrrr}
    \toprule
    \multirow{2}{*}{Method} & \multicolumn{1}{c}{\textbf{CIFAR-10}} & \multicolumn{1}{c}{\textbf{CIFAR-100}} & \multicolumn{1}{c}{\textbf{STL-10}} & \multicolumn{1}{c}{\textbf{LSUN}} & \multicolumn{1}{c}{\textbf{CelebA}} \\
    & sec ($\times 10$) & sec ($\times 10$) & sec ($\times 10$) & sec ($\times 10$) & sec ($\times 10$)  \\
    \midrule
    NSGAN  & {3.80}$\pm${0.04} & {3.72}$\pm${0.03} & {4.89}$\pm${0.06} & {10.93}$\pm${0.10} & {10.98}$\pm${0.10}  \\ %
    WGAN-GP  & {5.86}$\pm${0.46} & {6.12}$\pm${0.13} & {8.19}$\pm${0.18} & {18.78}$\pm${0.10}& {18.61}$\pm${0.10} \\ %
    SNGAN  & {4.17}$\pm${0.05} & {4.17}$\pm${0.04} & {5.57}$\pm${0.11} & {11.62}$\pm${0.11}& {11.56}$\pm${0.09}\\
    GraND-GAN (Ours)   & {5.66}$\pm${0.04} & {5.69}$\pm${0.04} & {8.83}$\pm${0.07} & {26.34}$\pm${0.08} & {26.13}$\pm${0.10}\\
    GraNC-GAN (Ours)  & {5.69}$\pm${0.04} & {5.65}$\pm${0.04} & {8.83}$\pm${0.08} & {26.11}$\pm${0.10}& {26.18}$\pm${0.18} \\ 
    \bottomrule
  \end{tabular}
  }
  
\end{table}

\begin{figure}
	\centering
    \begin{subfigure}[b]{0.22\textwidth}
		\includegraphics[width=1.0\textwidth]{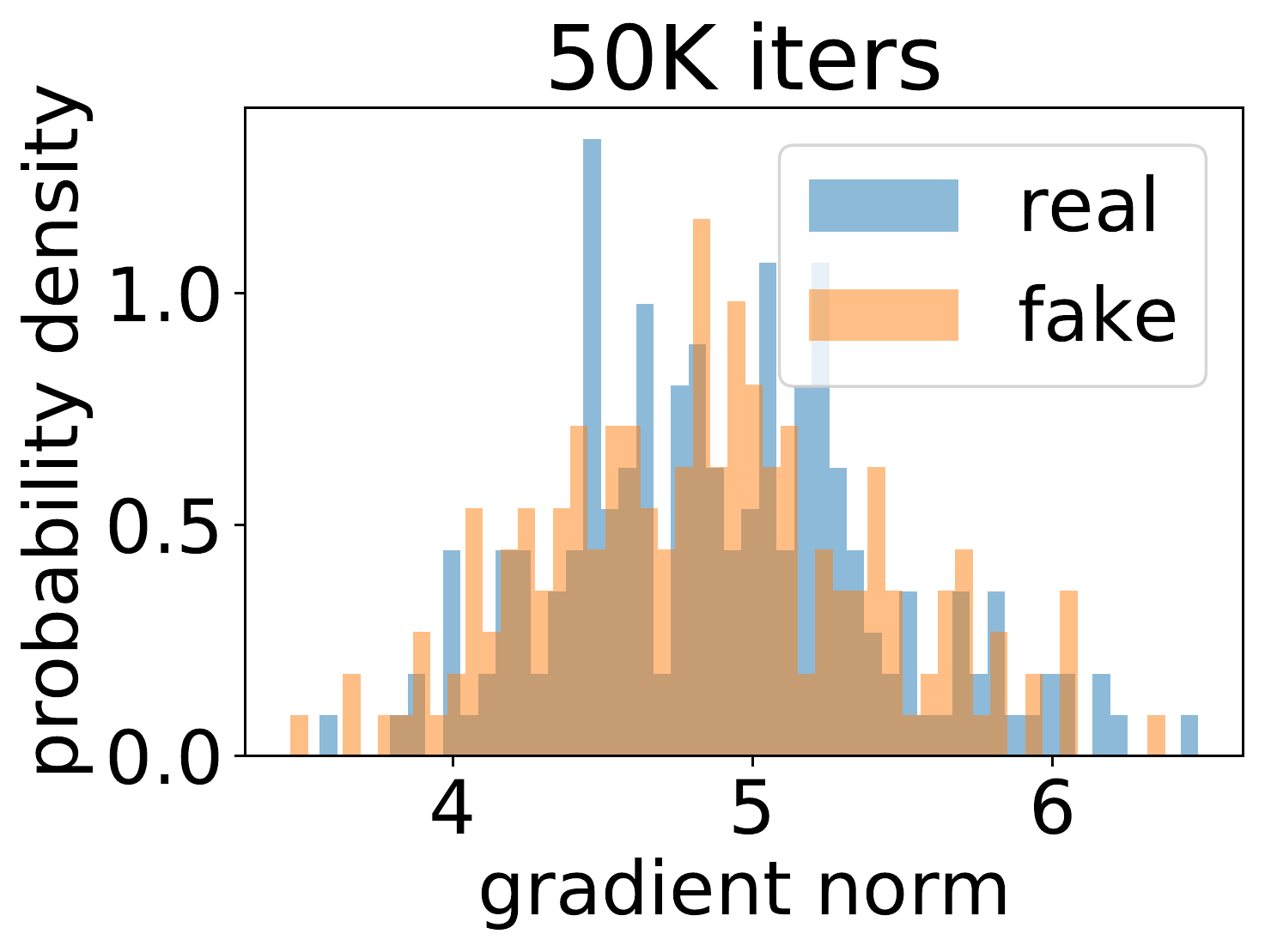}
        \caption{\label{fig:gradnormsa}Gradient norm (SNGAN with sum pooling)}
    \end{subfigure} \hfill
    \begin{subfigure}[b]{0.22\textwidth}
		\includegraphics[width=1.0\textwidth]{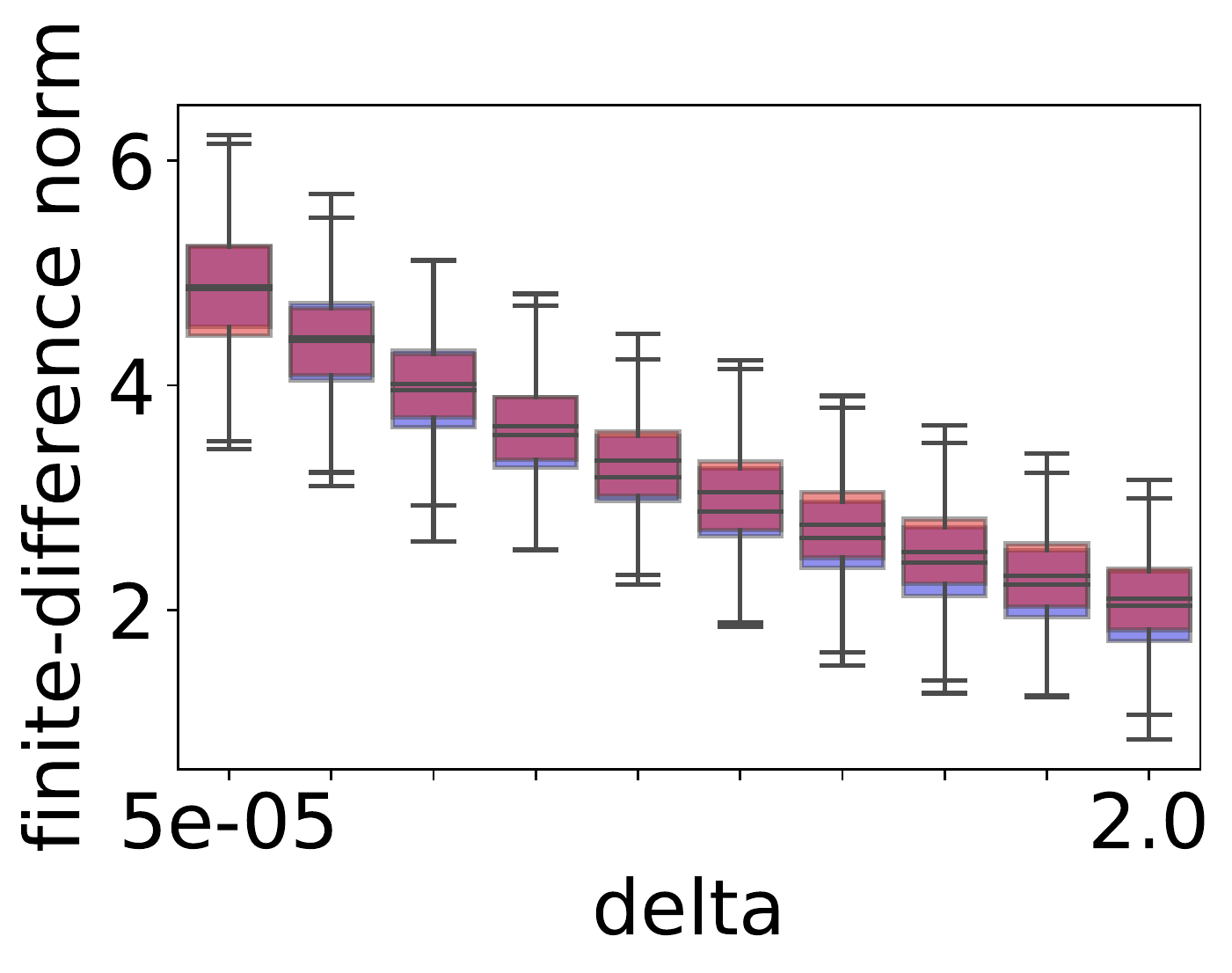}
        \caption{\label{fig:gradnormsb}FD gradient norm (SNGAN with sum pooling)}
    \end{subfigure}\\
    \begin{subfigure}[b]{0.22\textwidth}
		\includegraphics[width=1.0\textwidth]{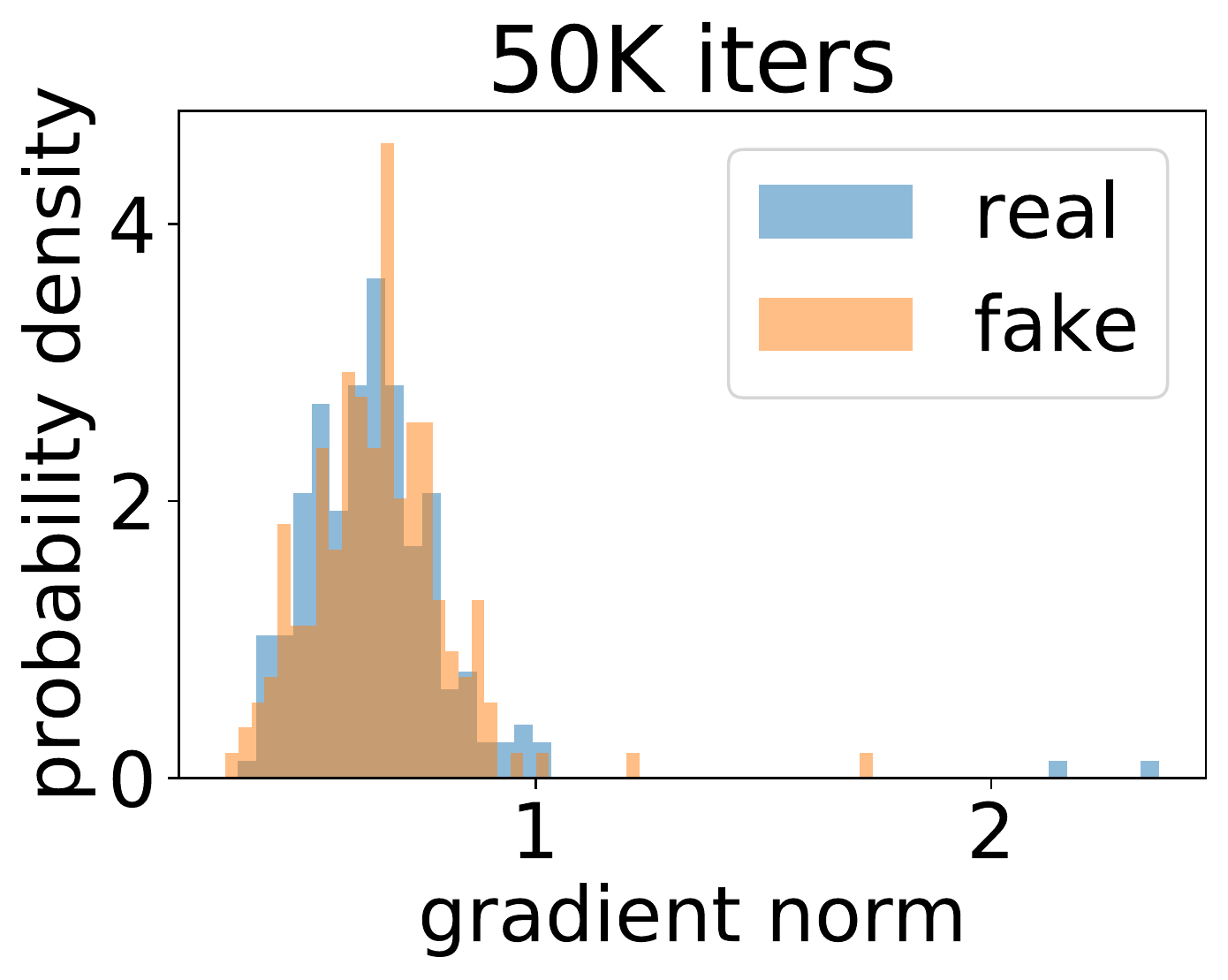}
        \caption{\label{fig:gradnormsc}Gradient norm (SNGAN with average pooling)}
    \end{subfigure}\hfill
    \begin{subfigure}[b]{0.22\textwidth}
		\includegraphics[width=1.0\textwidth]{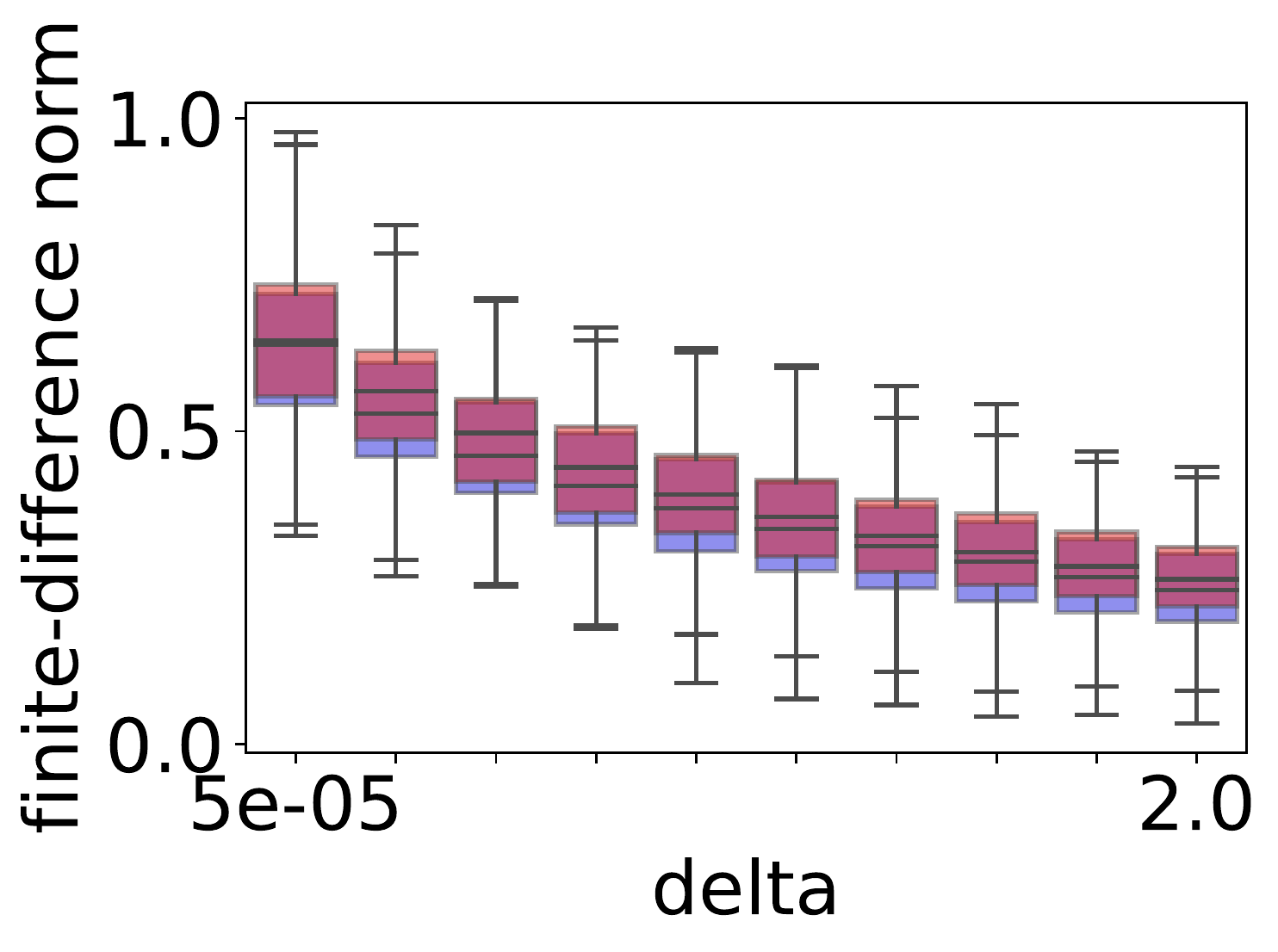}
        \caption{\label{fig:gradnormsd}FD gradient norm (SNGAN with average pooling)}
    \end{subfigure}
    \caption{\label{fig:gradnormsappendix} Gradient norms and finite-difference (FD) approximation to the gradient norms at increasing perturbation length $\texttt{delta}$ along the gradient for SNGAN with global sum pooling ((a) and (b)) and global average pooling ((c) and (d)) at 50K iterations of training on CIFAR-10.}
\end{figure}

This is because gradient normalized discriminator (or critic) $D$ requires computing the gradient norms on both the real and the fake samples when updating the parameters of $D$. In contrast, WGAN-GP only computes gradient norms on half the total number of real$+$fake samples which are random interpolates between the reals and fakes. Moreover, when updating $G$, computing generator loss $\mathcal{L}_G$ requires computing the gradient norm of $D$ for GraN models, unlike WGAN-GP where gradient penalty affects only the parameter updates for $D$ at a given training iteration. GraN and WGAN-GP are both slower relative to NSGAN or SNGAN because they involve computing the gradient norm and backpropagating through it. We remark that further advances in the efficiency of back-propagation through network gradients could ameliorate this issue (e.g., AutoInt \cite{Lindell_2021_CVPR}).

However, we note that, since the generator $G$ architecture is identical across methods for a given dataset, at inference all methods fare equally in wall-clock timings for image generation. 

\section{The Looseness of Layerwise Constraints} \label{appendixD}

We consider the simple case of the composition of two linear layers, $z = f(g(x)) = B (A x + a) + b$, where
both $f$ and $g$, have a sharp Lipschitz constant (LC) of one. The question is under what conditions does $f \circ g$ also have a sharp LC of one?

We first introduce some notation.  Let $M$ be an $m \times n$ matrix.  We denote the singular value decomposition of $M$ as $M = U_M \Sigma_M V_M^T$, where we take $U_M$ and $V_M$ to be square matrices (of sizes $m \times m$ and $n \times n$, respectively).  Here $\Sigma_M$ is a diagonal $m \times n$ matrix, with the non-negative singular values sorted in decreasing order down the diagonal \cite{golub2013matrix}.  Define $\sigma_1(M)$ to be the maximal SV of the matrix $M$.  Moreover, define $\Gamma_\sigma (M)$ to be the projection from $\mathbb{R}^n$ to the subspace spanned by the right singular vectors of $M$ for SV's equal to $\sigma$.  That is, 
\begin{equation}
\Gamma_\sigma (M) = V_M D_{\sigma}(M) V_M^T, \label{eq:GammaM}
\end{equation}
where $D_{\sigma}(M)$ is defined to be a diagonal $n \times n$ matrix
where the $i^{th}$ diagonal element is $1$ when the corresponding element of $\Sigma_M$ equals $\sigma$, and zero otherwise.  It then follows that $\Gamma_\sigma (M^T)$ is the projection of $\mathbb{R}^m$ to the subspace spanned by the {\em left} singular vectors of $M$ for the SV $\sigma$.  Finally, from the form of $\Gamma_\sigma(M)$ in (\ref{eq:GammaM}) we can conclude
\begin{align}
    & \sigma_1( \Gamma_\sigma(M) ) = 1, \label{eq:maxSigGamma} \\
    & \Gamma_\sigma(M) \Gamma_\sigma(M) = \Gamma_\sigma(M), \label{eq:projectionSigGamma} 
\end{align}
so long as $\sigma$ is an SV for $M$.

We can express the conditions that $f$, $g$, and $f \circ g$ all have sharp LC of one in terms of this notation.  Specifically, the tight Lipschitz bounds for $f$, $g$ and $f \circ g$ are $\sigma_1(B)$, $\sigma_1(A)$, and $\sigma_1(BA)$, respectively.  Assuming SN has rescaled $A$ and $B$ appropriately, then the LC of $g$ and $f$ are both one and we have $\sigma_1(B)$ $=$ $\sigma_1(A)$ $=$ $1$.  Moreover, we see $f \circ g$ will have a sharp LC of one iff $\sigma_1(BA)$ $=$ $1$.  We examine this latter condition.

\begin{theorem}
    For $A$ and $B$ as above (with dimensions such that their product $B A$ can be formed), with maximal SV's equal to one, the maximal SV of $B A$ satisfies $\sigma_1(BA) \leq 1$.
    Further, equality of this bound holds if and only if 
    \begin{equation}
       \sigma_1(\Gamma_1(B) \Gamma_1(A^T)) = 1 . \label{eq:subspaceAlign}
    \end{equation}
\end{theorem}

\noindent
{\bf Proof of Theorem 1.}
Let $m_1 = \text{dim}(\Gamma_1(A))$ and $n_1 = \text{dim}(\Gamma_1(B))$ be the number of singular values equal to one in $A$ and $B$, respectively.  The assumption that $\sigma_1(A) = \sigma_1(B) = 1$ implies $m_1, n_1 > 0$.
    
Recall that the spectral norm of a matrix $M$, which is induced by the $L_2$ vector norm, can be defined via the largest singular value:
$|| M ||_2 := \sigma_1(M)$,
equivalently computed as
\begin{equation}
        ||M||_2 
        = \sup_{||x|| = ||y|| = 1} |y^T M x| 
        = \sup_{x\neq 0} \frac{||M x ||}{||x||}. \label{eq:maxSV}
\end{equation}
It follows from (\ref{eq:maxSV}) that $|| B A ||_2 \leq || B ||_2 ||A||_2$ (the submultiplicativity property), and hence
\begin{equation}
    \sigma_1(BA) \leq \sigma_1(B) \sigma_1(A) = 1.  \label{eq:bndSigmaBA}
\end{equation}
 
First we prove $\sigma_1(BA) = 1$ implies (\ref{eq:subspaceAlign}).  Assume $\sigma_1(BA) = 1$. Then it follows from (\ref{eq:maxSV}) that there exists a vector $x$ such that  $|| x || = 1$ and $z := B A x$ satisfies $|| z || = 1$.  Let $y = A x$.  There are two cases to consider, either $|| y || < 1$, or $|| y || = 1$.  However, since $z = B y$ and $|| B||_2 = \sigma_1(B) = 1$ we have $|| z || <= || B ||_2 ||y|| = ||y||$.  Therefore the assumption $||y || < 1$ leads to the contradiction $|| z || < 1$, and instead we must have
\begin{align}
    ||y|| &= ||A x || = || x || = 1, \label{eq:Ax} \\
   ||z|| &= ||B y || = || y || = 1. \label{eq:By}
\end{align}
Equation (\ref{eq:Ax}) ensures $x \in \mbox{\tt range}(\Gamma_1(A))$ and therefore $y = Ax \in \mbox{\tt range}(\Gamma_1(A^T))$.  Also, equation (\ref{eq:By}) ensures $y \in \mbox{\tt range}(\Gamma_1(B))$.  Therefore it follows that $y$ is a unit vector such that $\Gamma_1(B) y = y$, and $\Gamma_1(A^T) y = y$.  And thus, $ \Gamma_1(B)\Gamma_1(A^T) y = y$. By (\ref{eq:maxSV}) we then have $\sigma_1(\Gamma_1(B)\Gamma_1(A^T)) \geq 1$.  But, from (\ref{eq:maxSigGamma}), it follows that $\sigma_1(\Gamma_1(B)) = \sigma_1(\Gamma_1(A^T)) = 1$ and therefore $\sigma_1(\Gamma_1(B)\Gamma_1(A^T)) \leq 1$. As a result we have shown (\ref{eq:subspaceAlign}), as desired.

For the reverse direction, assume $\sigma_1(\Gamma_1(B)\Gamma_1(A^T)) = 1$.  Then (\ref{eq:maxSV}) implies there exists a $y$ such that $|| y || = 1$ and $z = \Gamma_1(B)\Gamma_1(A^T) y$ with $||z|| = 1$.  But, since $\Gamma_1(B)$ and $\Gamma_1(A^T)$ are projection matrices (see (\ref{eq:projectionSigGamma})), it can be shown that we must have $\Gamma_1(A^T) y = y$, $\Gamma_1(B) y = y$, and $z = y$.\footnote{The basic idea here is that if $z = P y$ for a projection $P$ and $||z || = || y||$ then $ ||y||^2 = ||Py||^2 + ||(I-P)y||^2$  can be used to show $(I-P) y = 0$. Moreover, from (\ref{eq:GammaM}), it then follows that $Py = y$.} 
Moreover, since $y$ is a right singular vector of $A^T$ for singular value one, it follows that $x := A^T y = V_A \Sigma_A U_A^T y$ is a left singular vector of $A^T$ for the SV at one and $|| x || = 1$. Therefore $x = \Gamma_1(A)x = V_A D_1(A) V_A^T x$. That is, $x$ is in the right singular space of $A$ for the SV at one and it follows that $y = A x$.  Taken together, we have $BA x$ $=$ $B y$ and $y = \Gamma_1(B) y$, so $||B y|| = 1$. Hence $||BA x|| = || B y || = 1 = ||x||$ .  That is, from (\ref{eq:maxSV}), it follows that $\sigma_1(BA) \geq 1$.  Finally, from (\ref{eq:bndSigmaBA}), we conclude $\sigma_1(BA) = 1$, as desired. $\square$

\textbf{Relation to Layerwise Spectral Normalization}~~%
As described in the text, (Eq.\ \ref{eq:subspaceAlign}) is a subspace alignment condition where $\sigma_1(\Gamma_1(B) \Gamma_1(A^T))$ equals the cosine of the first principal angle between the two subspaces $\Gamma_1(B)$ and $\Gamma_1(A^T)$ \cite{golub2013matrix,zhu2012angles}.  It is therefore unlikely to be satisfied by chance, although during training the model may reduce this angle and approach 
$\sigma_1(\Gamma_1(B) \Gamma_1(A^T)) = 1$. Thus, with training, we might expect the norm of the gradients of $f(x)$ to increase towards an upper bound.

However, we note additional features of $f(x)$ for the architectures described in Figures \ref{fig:discarch} and \ref{fig:blockarchs}.  Specifically, the global sum pooling in Fig.\ \ref{fig:discarch} and the skip connections in Fig. \ref{fig:blockarchs} are both capable of amplifying the gradient norms through these stages by a factor greater than one.  Thus, while the subspace alignment conditions can be expected to shrink the gradient magnitudes, these specific components can expand them. The consequence of these two opposing effects is not clear a priori.  

The empirical results shown in Fig.\ \ref{fig:gradnormsa} indicate that, for the cases tested, the net effect is for $f(x)$ to have a gradient norm larger than one. Moreover, when average pooling is used in place of the sum pooling, the norm of the gradient is predominantly less than one (see Fig.\ \ref{fig:gradnormsc}). Similar properties are seen for the magnitudes of finite differences (FD) of $f(x)$ over steps of length $\delta$ (see Fig.\ \ref{fig:gradnormsb}, \ref{fig:gradnormsd}), as described in the paper.

Indeed, we can compute the LC for the $32\times 32$ resnet-based convolutional discriminator (used on, e.g., CIFAR-10), shown in \S\hyperref[appendixB]{B} and Fig.\ \ref{fig:discarch}, as follows.
First, note that the four \verb|DBlocks| have a skip connection, meaning the LC increases two-fold across each block, resulting in an LC of $2^4 = 16$ before pooling (assuming the SN keeps the convolutional layer LCs at one). 
The first two blocks also have spatial downsampling, resulting in an $8\times 8$ feature map that is sum-pooled. 
This pooling, along with the preceding skip connections, increases the final LC to $8\times 8 \times 16 = 1024$, as mentioned in the main paper.

\begin{table}[t]
\centering
  \caption{\label{tab:variability} Mean $\pm$ standard deviation of IS, FID and KID across 3 training runs with random restarts on CIFAR-10. $\dagger$ indicates modified baselines with the Lipschitz constant $\mathcal{K}=0.83$ that our methods use. 
  }
  \small{
  \begin{tabular}[t]{lrrr}
    \toprule
    \multirow{1}{*}{Model} & \multicolumn{1}{c}{\textbf{IS}~$\uparrow$} & \multicolumn{1}{c}{\textbf{FID}~$\downarrow$} & \multicolumn{1}{c}{\textbf{KID}($\times1000$)~$\downarrow$} \\
    \midrule
    NSGAN & {7.35}$\pm${0.25} & {26.85}$\pm${5.16} & {17.81}$\pm${3.79}  \\
    WGAN-GP & {7.42}$\pm${0.02} & {22.44}$\pm${0.35} & {20.67}$\pm${0.31}  \\
    SNGAN & {8.06}$\pm${0.04} & {17.22}$\pm${0.16} & {12.44}$\pm${0.25}  \\
    \midrule
    NSGAN-GP$\dagger$ & {8.01}$\pm${0.04} & {15.69}{$\pm$}{0.15} & {12.95}$\pm${0.21}  \\
    NSGAN-SN$\dagger$ & {7.72}$\pm${0.06} & {21.12}$\pm${0.59} & {15.79}$\pm${0.42}  \\
    WGAN-GP$\dagger$ & {7.37}$\pm${0.02} & {22.75}$\pm${0.05} & {21.12}$\pm${0.36}  \\
    SNGAN$\dagger$ & {7.98}$\pm${0.01} & {16.86}$\pm${0.40} & {12.16}$\pm${0.38}  \\
    \midrule
    GraND-GAN & {8.00}$\pm${0.01} & {15.60}{$\pm$}{0.47} & {12.80}$\pm${0.42} \\
    GraNC-GAN & {7.96}$\pm${0.02} & {16.15}$\pm${0.21} & {13.30}$\pm${0.32}\\
    \bottomrule
  \end{tabular}
  }
\end{table}

\begin{table}[t]
\centering
  \caption{\label{tab:ablation-lipschitz} Ablation of our method (GraND-GAN) on CIFAR-10 image generation under different values of the Lipschitz constant $\mathcal{K}=1/\tau$ with $\epsilon=0.1$. %
  }
  \small{
  \begin{tabular}[t]{lrrr}
    \toprule
    \multirow{1}{*}{$1/\tau$} & \multicolumn{1}{c}{\textbf{IS}~$\uparrow$} & \multicolumn{1}{c}{\textbf{FID}~$\downarrow$} & \multicolumn{1}{c}{\textbf{KID}($\times$1000)~$\downarrow$} \\
    \midrule
    0.1 & {7.709} & {18.303} & {15.4}  \\
    0.5 & {7.919} & {15.689} & {12.8}  \\
    0.83 & {8.031} & {14.965} & {12.3}  \\
    1.0 & {8.011} & {15.469} & {12.2}  \\
    1.33 & {8.111} & {14.561} & {10.9}  \\
    \bottomrule
  \end{tabular}
  }
\end{table}

\begin{table}[t]
\centering
  \caption{\label{tab:ablation-epsilons} Ablation of our method (GraND-GAN) on CIFAR-10 image generation under different values of $\epsilon$ with $\mathcal{K}=1/\tau=0.83$. %
  }
  \small{
  \begin{tabular}[t]{lrrr}
    \toprule
    \multirow{1}{*}{$\epsilon$} & \multicolumn{1}{c}{\textbf{IS}~$\uparrow$} & \multicolumn{1}{c}{\textbf{FID}~$\downarrow$} & \multicolumn{1}{c}{\textbf{KID}($\times$1000)~$\downarrow$} \\
    \midrule
    1e-08 & {8.065} & {15.076} & {11.9}  \\
    0.0001 & {7.924} & {16.695} & {13.7}  \\
    0.001 & {8.035} & {16.322} & {13.5}  \\
    0.01 & {7.900} & {15.726} & {13.0}  \\
    0.1 & {8.031} & {14.965} & {12.3}  \\
    1.0 & {7.981} & {15.194} & {12.2}  \\
    \bottomrule
  \end{tabular}
  }
\end{table}

\begin{table}[t]
  \caption{\label{tab:diverge}
  Frequency of runs diverging (i.e., $\text{FID}\geq40$) on CIFAR-10 on three random restarts for GraND-GAN, GraNC-GAN, SNGAN, WGAN-GP, and NSGAN-GP$\dagger$ on CIFAR-10 for settings B, C and D.}
  \vspace{1mm}
  \centering
   \resizebox{0.47\textwidth}{!}{
  \begin{tabular}{lrrrrrr}
    \toprule
    {Model} & {Setting} & $\alpha$ & $\beta_1$ & $\beta_2$ & $n_{\rm dis}$ & $\#(\text{FID}\geq40)$\\
    \midrule
    WGAN-GP & B & 0.0002 & 0.5 & 0.999 & 1 & 3/3 \\
    & C & 0.001 & 0.5 & 0.999 & 5 & 3/3 \\
    & D & 0.001 & 0.9 & 0.999 & 5 & 3/3 \\
    \midrule
    SNGAN & B & 0.0002 & 0.5 & 0.999 & 1 & 3/3 \\
    & C & 0.001 & 0.5 & 0.999 & 5 & 0/3 \\
    & D & 0.001 & 0.9 & 0.999 & 5 &  0/3 \\
    \midrule
    NSGAN-GP$\dagger$ & B & 0.0002 & 0.5 & 0.999 & 1 & 3/3 \\
    & C & 0.001 & 0.5 & 0.999 & 5 & 3/3 \\
    & D & 0.001 & 0.9 & 0.999 & 5 &  0/3 \\
    \midrule
    GraND-GAN & B & 0.0002 & 0.5 & 0.999 & 1 & 2/3 \\
    & C & 0.001 & 0.5 & 0.999 & 5 & 0/3 \\
    & D & 0.001 & 0.9 & 0.999 & 5 & 0/3 \\
    \midrule
    GraNC-GAN & B & 0.0002 & 0.5 & 0.999 & 1 & 2/3 \\
    & C & 0.001 & 0.5 & 0.999 & 5 & 0/3 \\
    & D & 0.001 & 0.9 & 0.999 & 5 & 0/3 \\
  \bottomrule
  \end{tabular}
  }
\end{table} 

\section{Variance of IS, FID and KID metrics across random training restarts for CIFAR-10}\label{appendixE}
We report the mean and the standard deviations of the metrics reported (IS, FID, KID) across 3 different training runs with random restarts for CIFAR-10 in Table \ref{tab:variability}.

\section{Ablations on $\epsilon$ and $\tau$}

We also run ablations on our methods by varying the piecewise Lipschitz constant $\mathcal{K}=1/\tau$  and $\epsilon$ for GraND-GAN on CIFAR-10 image generation. Tables \ref{tab:ablation-lipschitz} and \ref{tab:ablation-epsilons}  show that our method is fairly robust to a range of $\mathcal{K}$ and $\epsilon$, respectively, on CIFAR-10. The role of hyperparameter $\epsilon$ is mainly numerical stability when the gradient norm becomes vanishingly small. Irrespective of $\epsilon$ used, we empirically find that the weights of the network scale up sufficiently large such that the input gradient norm of the GraNed output $g(x)$ is close to the upper bound $\mathcal{K}$, i.e., $\|\nabla_x g(x)\|\approx \mathcal{K}$. This is evident in Figure 3 of the main paper where the gradient norms for our methods have a very narrow distribution around $\mathcal{K}$ despite using $\epsilon=0.1$.

\begin{figure}[ht]
	\centering 
    \begin{subfigure}[b]{0.23\textwidth}
		\includegraphics[width=1.0\textwidth]{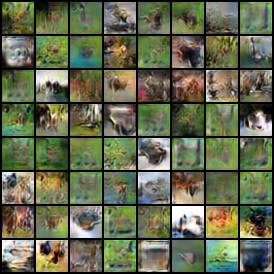}
        \caption{$\epsilon_{\text{Adam}}=1\times10^{-8}$}
    \end{subfigure} \hfill
    \begin{subfigure}[b]{0.23\textwidth}
		\includegraphics[width=1.0\textwidth]{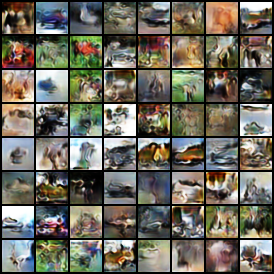}
        \caption{$\epsilon_{\text{Adam}}=1\times10^{-7}$}
    \end{subfigure}\hfill
    \caption{\label{fig:epsadam} Qualitative comparison of generated CIFAR-10 samples under two different $\epsilon_{\text{Adam}}$ hyperparameter settings. Tuning $\epsilon_{\text{Adam}}$ affects GAN training. Zoom in for better viewing.}
\end{figure}

\section{Frequency of runs diverging on CIFAR-10 on three random restarts}\label{appendixG}
We repeat the experiment in Figure 2 of the main paper for settings B ($\alpha=0.0002,~\beta_1=0.5,~\beta_2=0.999,~n_{\text{dis}}=1$), C ($\alpha=0.001,~\beta_1=0.5,~\beta_2=0.999,~n_{\text{dis}}=5$) and D ($\alpha=0.001,~\beta_1=0.9,~\beta_2=0.999,~n_{\text{dis}}=5$) that have aggressive training settings over 3 random restarts. We call a run ``diverging" when the FID $\geq 40$ on CIFAR-10. Table \ref{tab:diverge} summarizes the number of diverging runs out of 3 random restarts.  

Evidently, GraND-GAN and GraNC-GAN have the least number of runs that diverged in 3 random restarts across settings B, C and D. SNGAN comes close but diverges $3/3$ times for setting B when $n_{\text{dis}}=1$. WGAN-GP performs poorly across random restarts for B, C and D. NSGAN-GP$\dagger$ diverges $3/3$ times for settings B and C.

\section{Qualitative Results}\label{appendixH}

\begin{figure*}
	\centering 
    \begin{subfigure}[b]{0.33\textwidth}
		\includegraphics[width=1.0\textwidth]{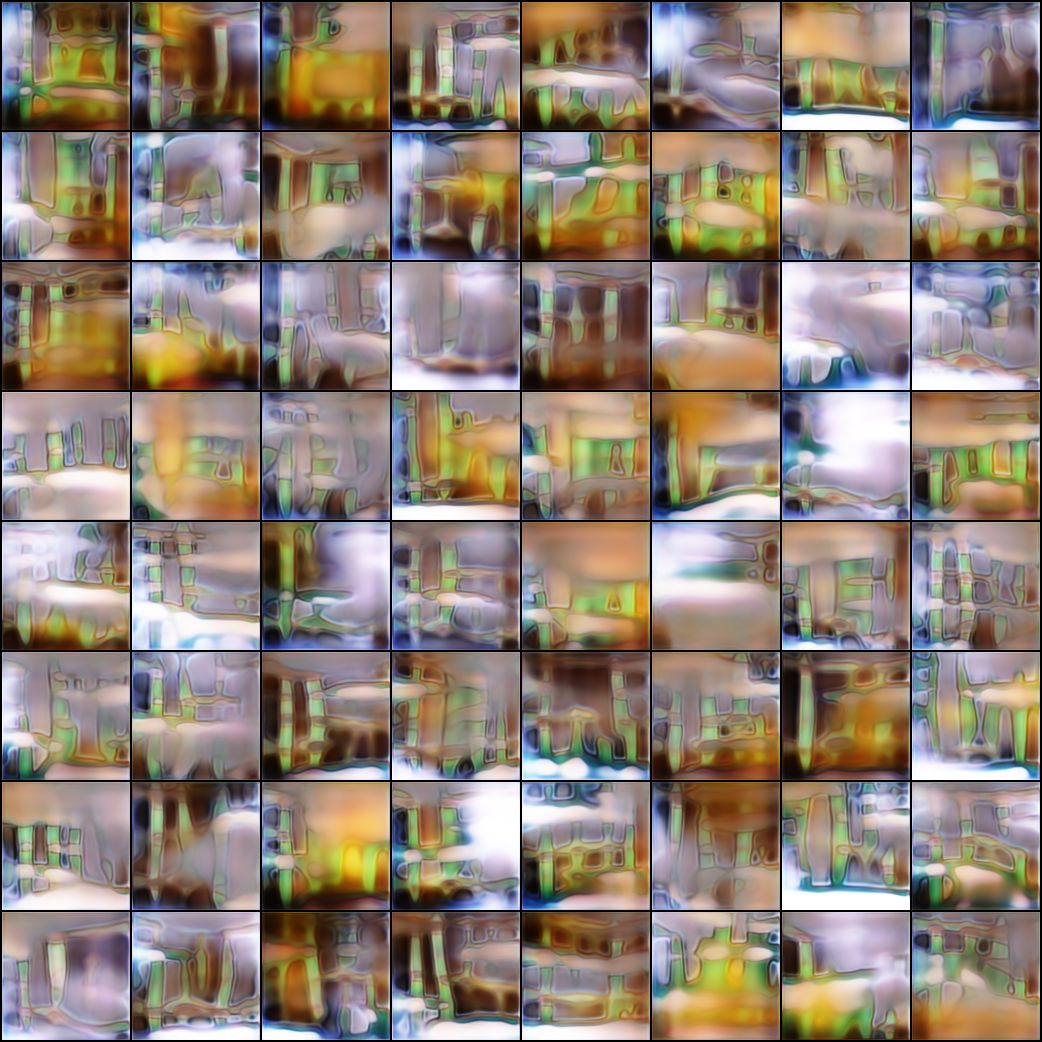}
        \caption{NSGAN}
    \end{subfigure} \hfill
    \begin{subfigure}[b]{0.33\textwidth}
		\includegraphics[width=1.0\textwidth]{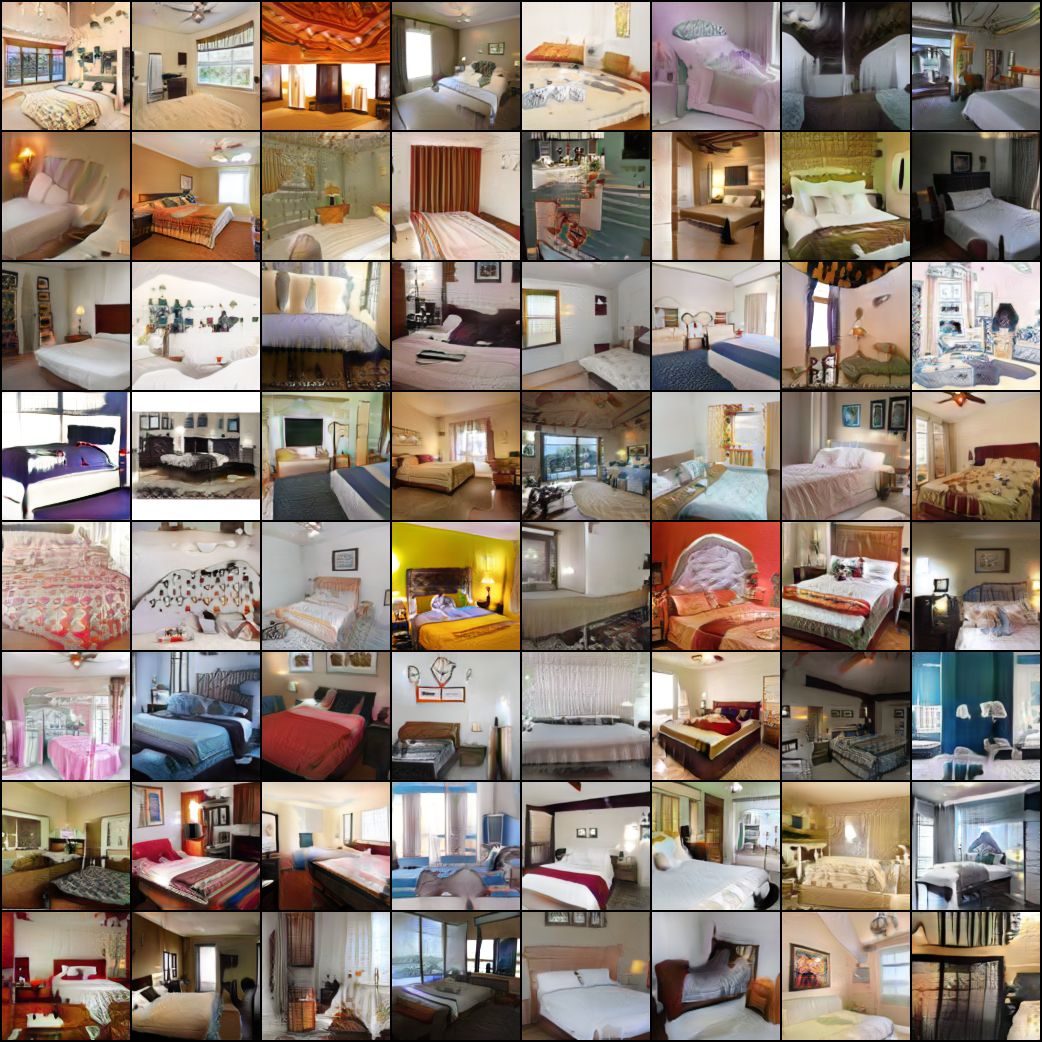}
        \caption{WGAN-GP}
    \end{subfigure}\hfill
    \begin{subfigure}[b]{0.33\textwidth}
		\includegraphics[width=1.0\textwidth]{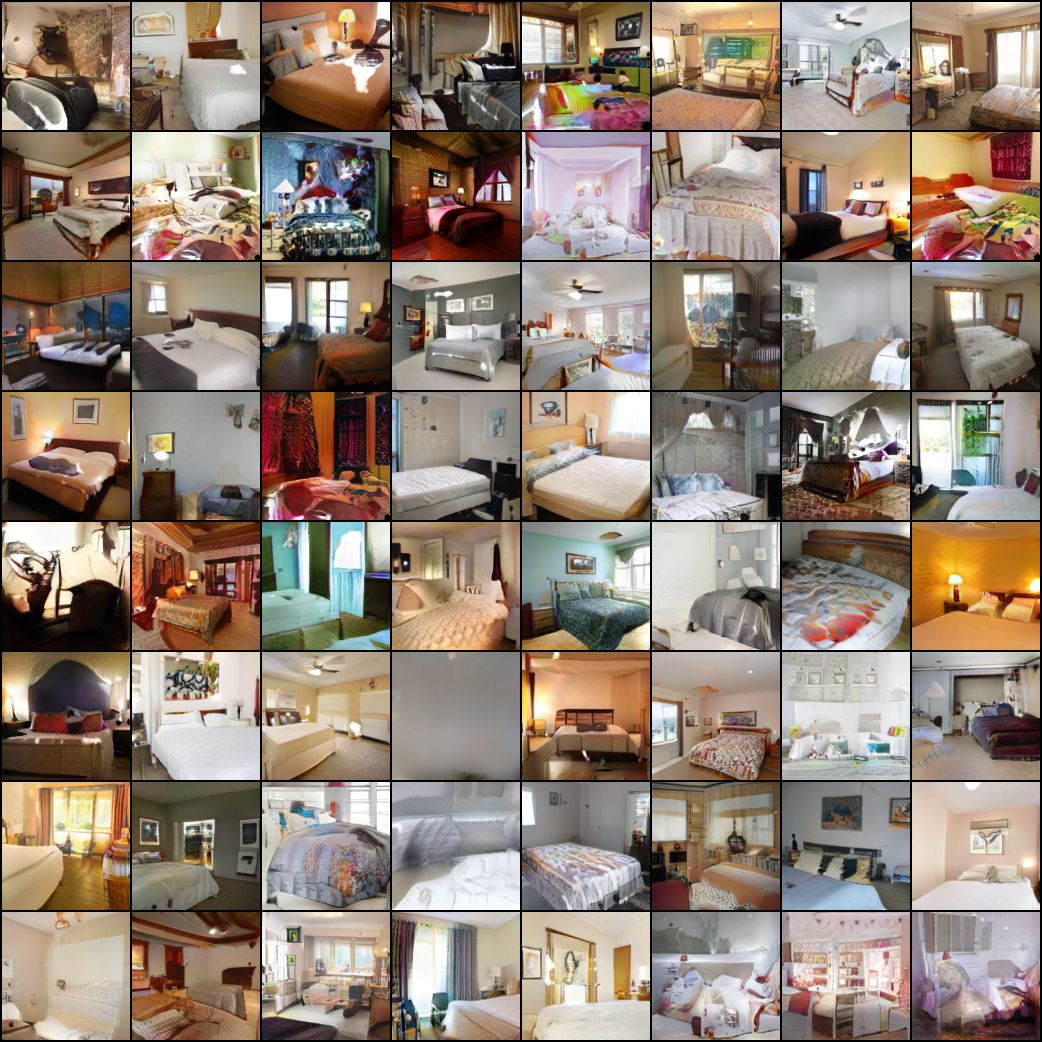}
        \caption{SNGAN}
    \end{subfigure}\\
    \begin{subfigure}[b]{0.33\textwidth}
		\includegraphics[width=1.0\textwidth]{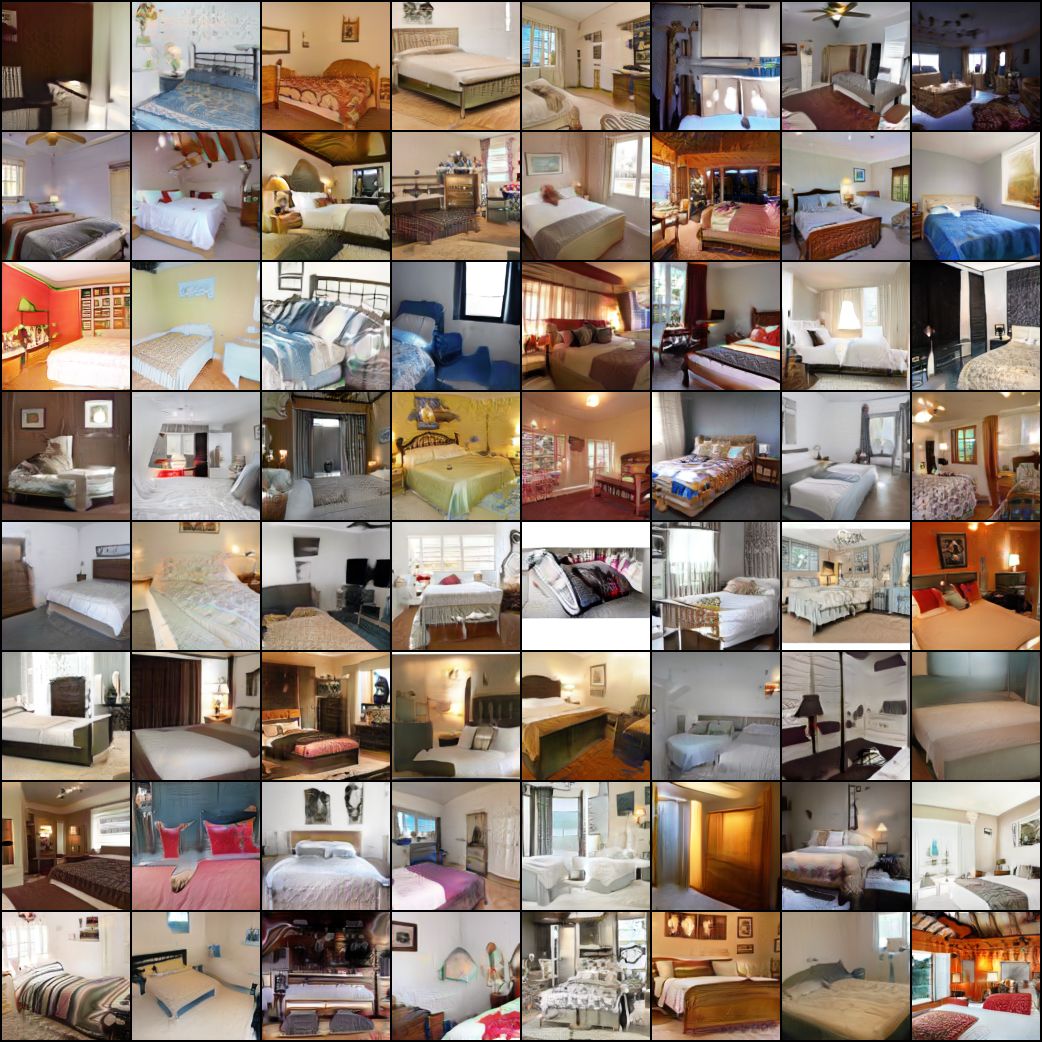}
        \caption{NSGAN-GP$\dagger$}
    \end{subfigure}\hfill
    \begin{subfigure}[b]{0.33\textwidth}
		\includegraphics[width=1.0\textwidth]{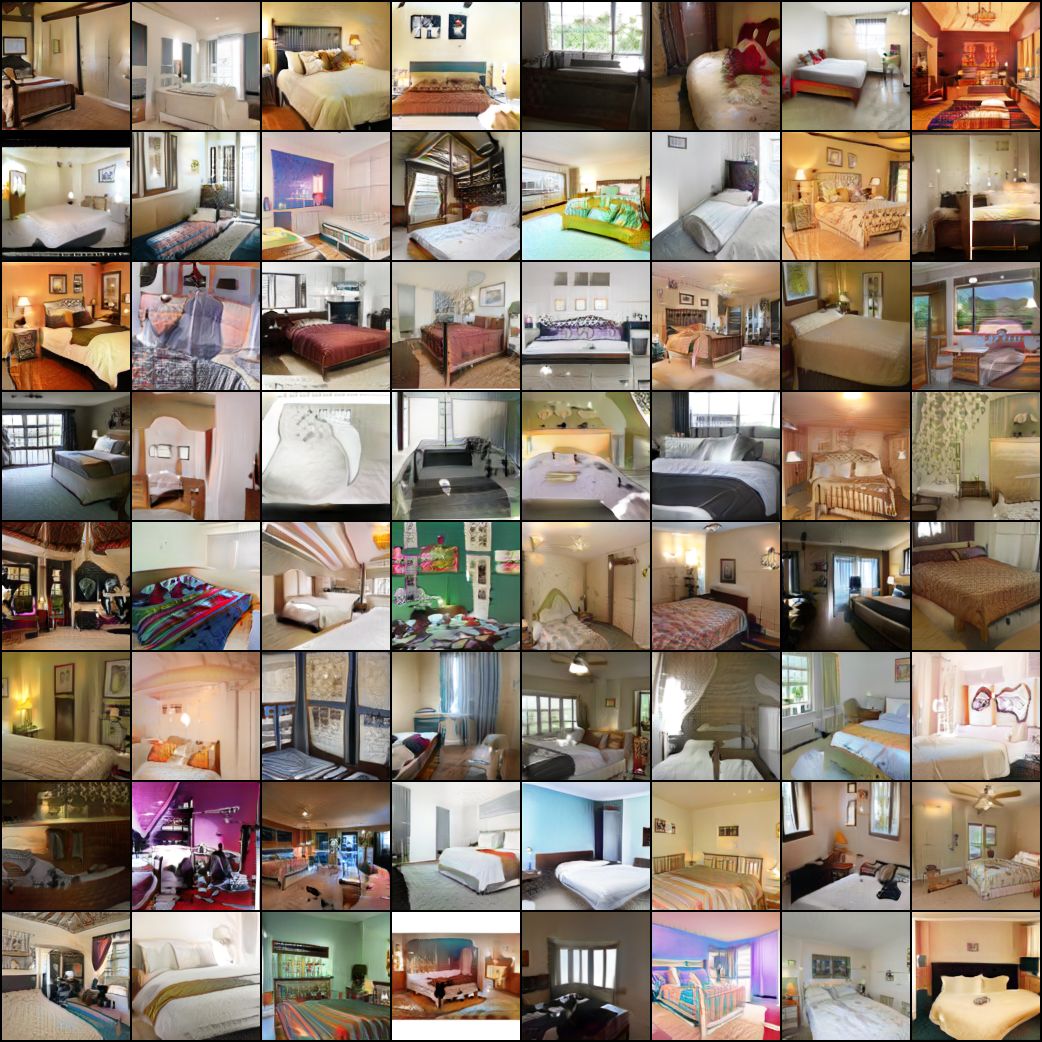}
        \caption{GraND-GAN}
    \end{subfigure}\hfill
        \begin{subfigure}[b]{0.33\textwidth}
		\includegraphics[width=1.0\textwidth]{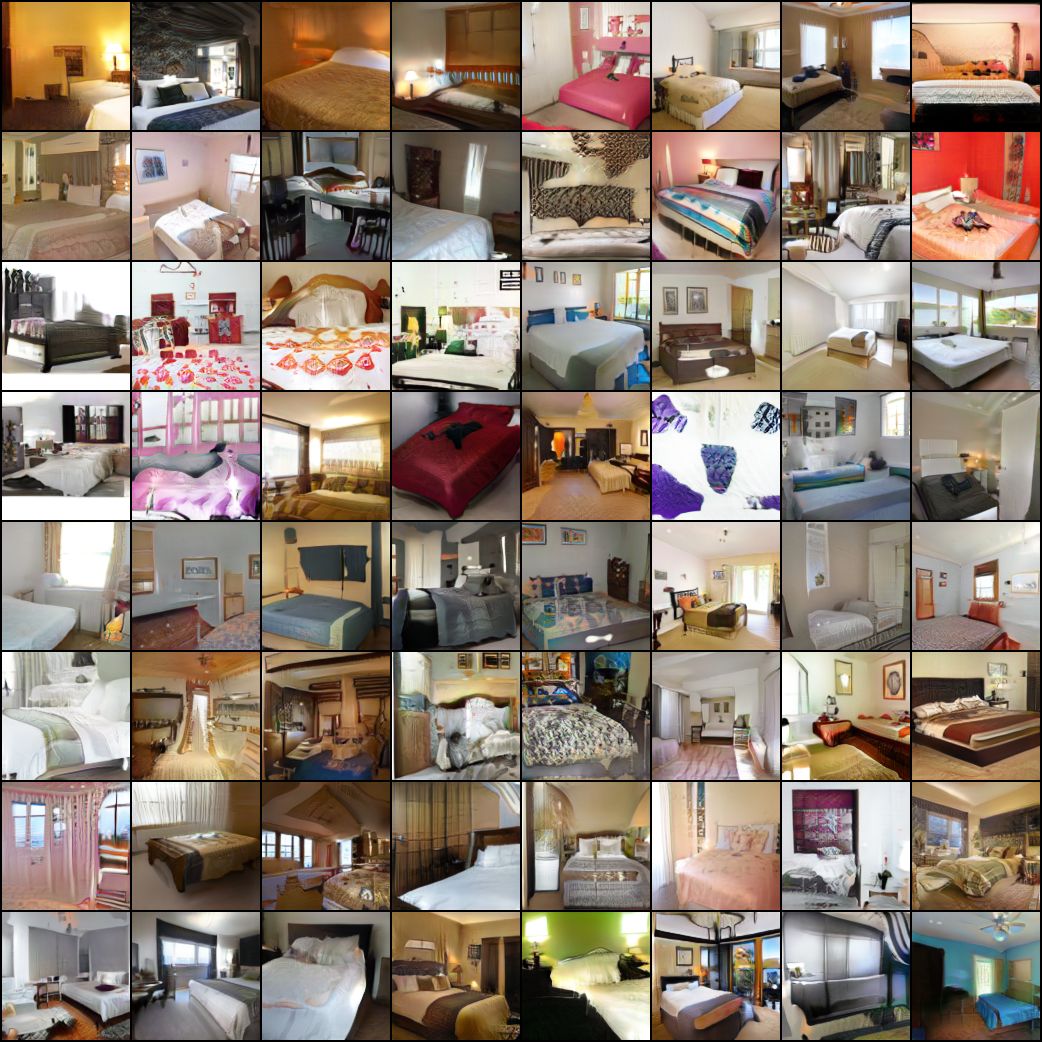}
        \caption{GraNC-GAN}
    \end{subfigure}
    \caption{\label{fig:qualitative-lsun} Qualitative results on LSUN-Bedrooms across different models, including baselines (NSGAN, WGAN-GP, SNGAN), the best performing modified baseline (NSGAN-GP$\dagger$) and our methods (GraND-GAN and GraNC-GAN). Zoom in for better viewing.}
\end{figure*}

\begin{figure*}
	\centering 
    \begin{subfigure}[b]{0.33\textwidth}
		\includegraphics[width=1.0\textwidth]{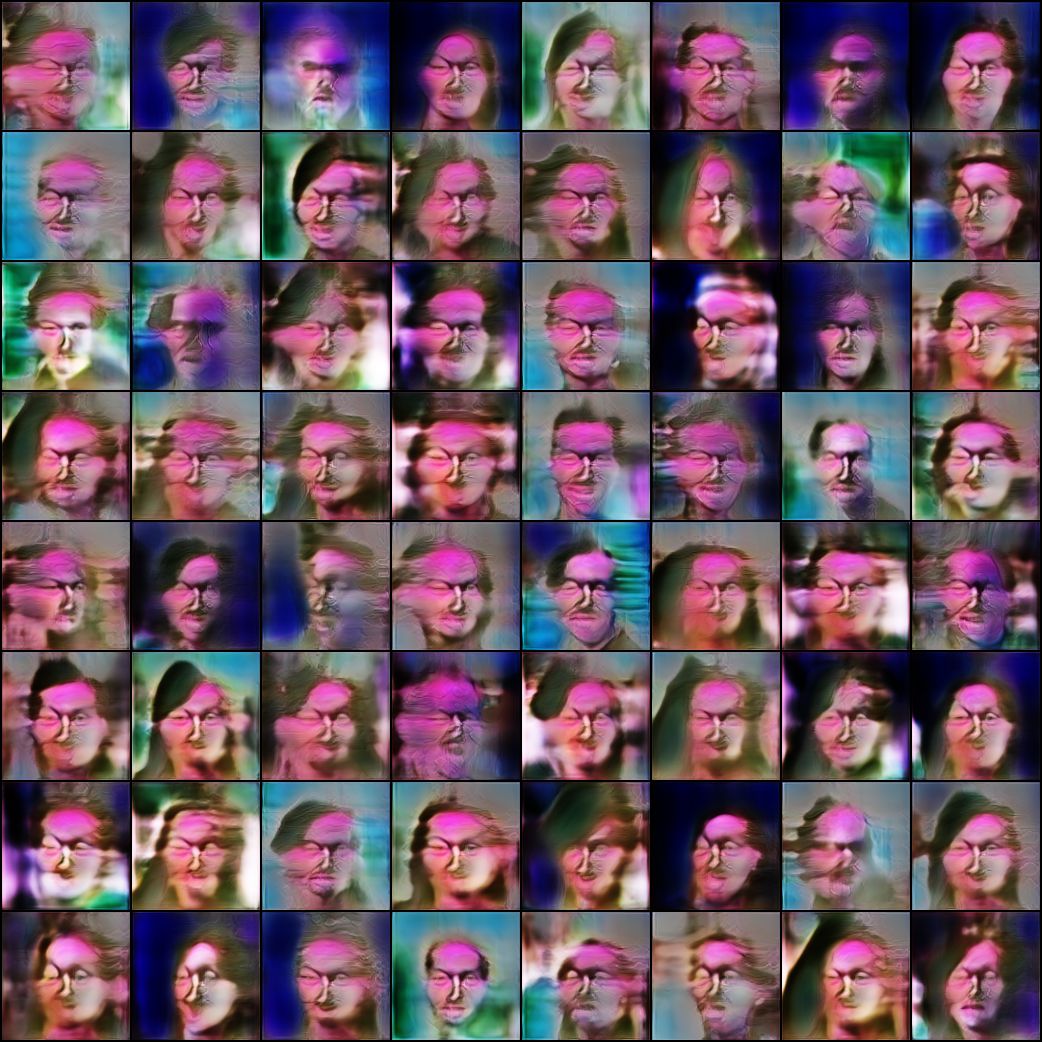}
        \caption{NSGAN}
    \end{subfigure} \hfill
    \begin{subfigure}[b]{0.33\textwidth}
		\includegraphics[width=1.0\textwidth]{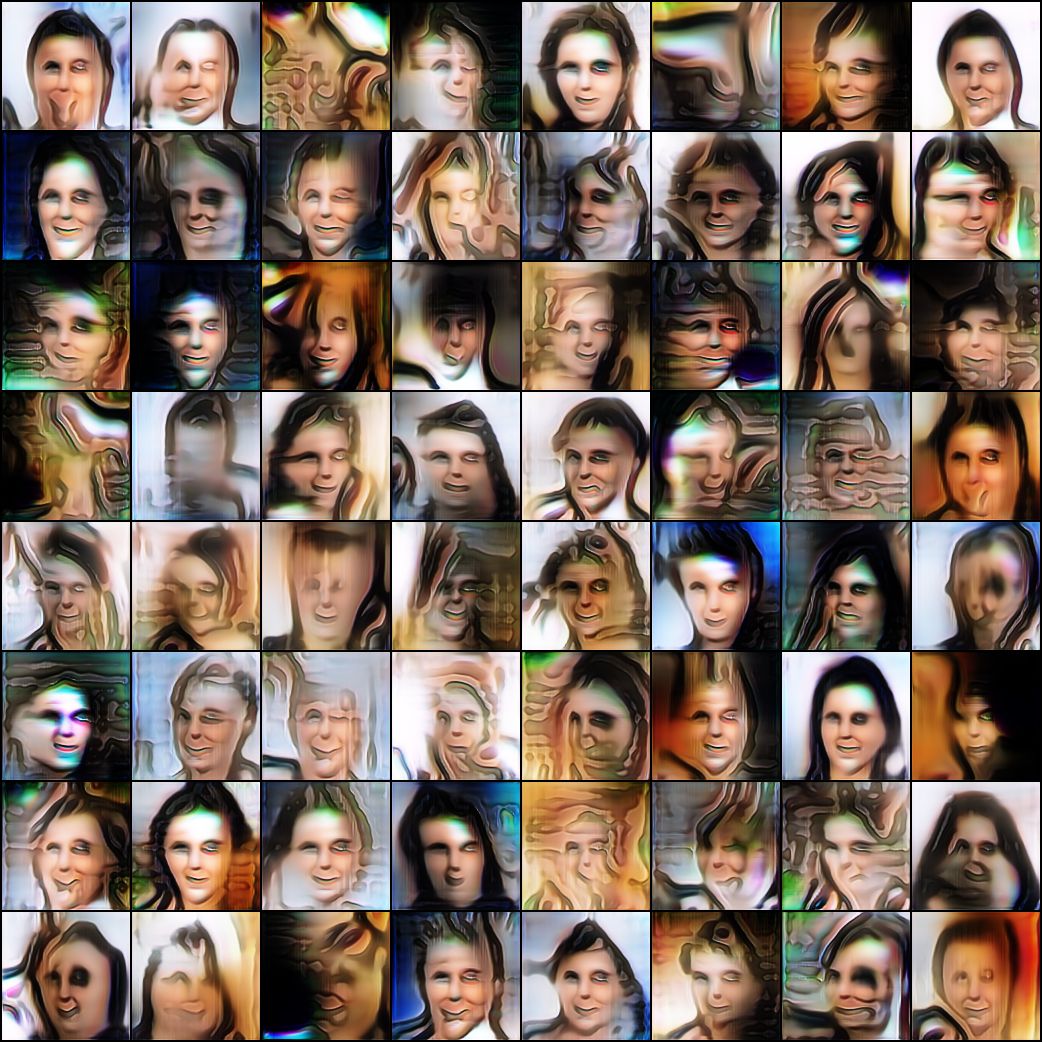}
        \caption{WGAN-GP}
    \end{subfigure}\hfill
    \begin{subfigure}[b]{0.33\textwidth}
		\includegraphics[width=1.0\textwidth]{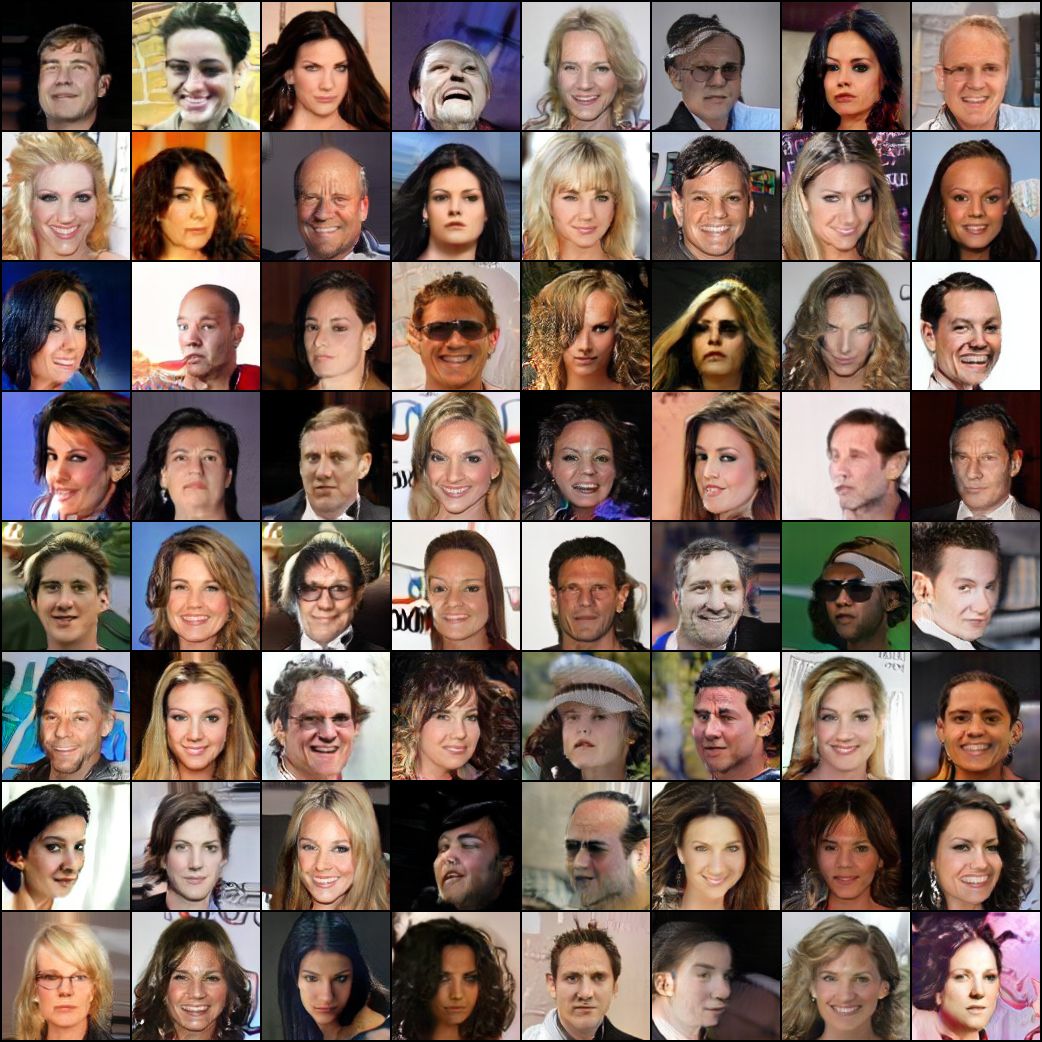}
        \caption{SNGAN}
    \end{subfigure}\\
    \begin{subfigure}[b]{0.33\textwidth}
		\includegraphics[width=1.0\textwidth]{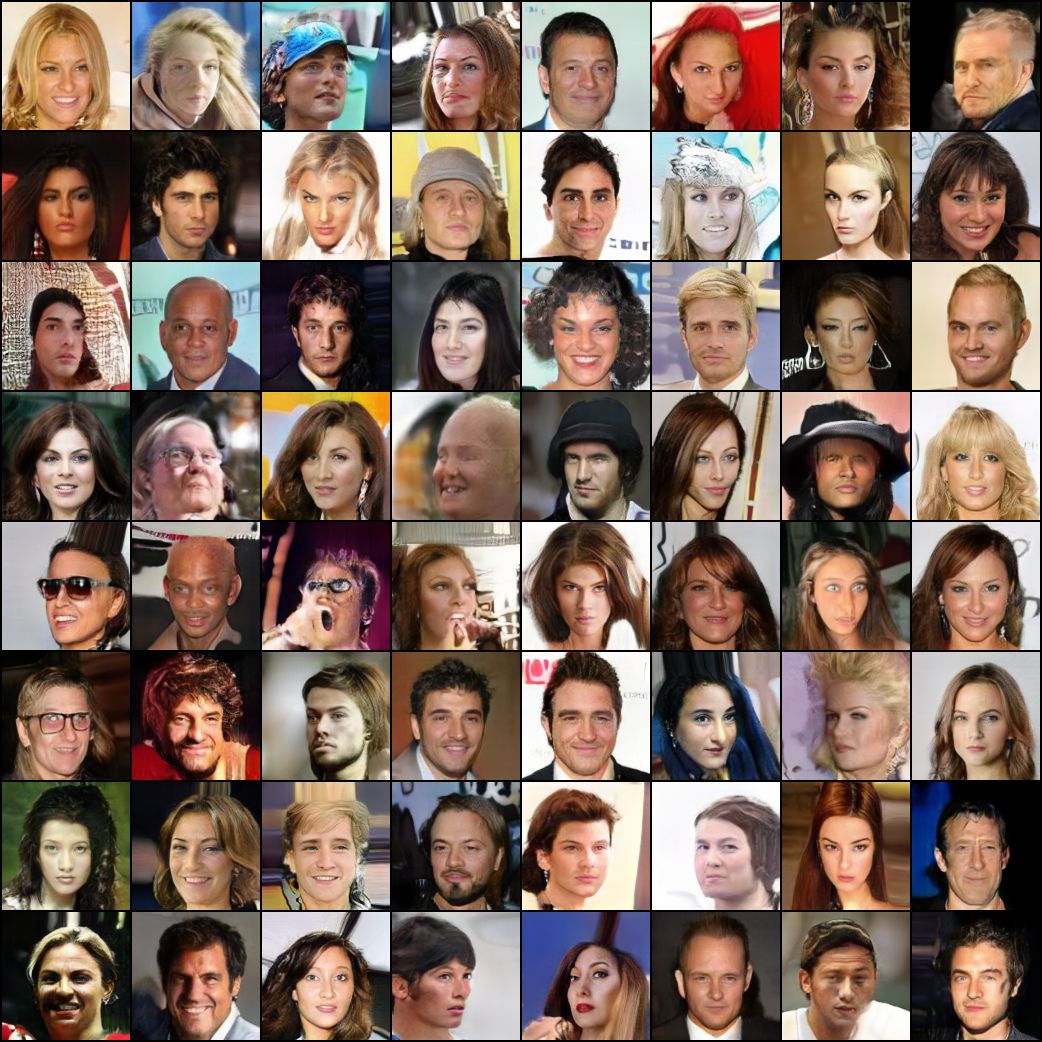}
        \caption{NSGAN-GP$\dagger$}
    \end{subfigure}\hfill
    \begin{subfigure}[b]{0.33\textwidth}
		\includegraphics[width=1.0\textwidth]{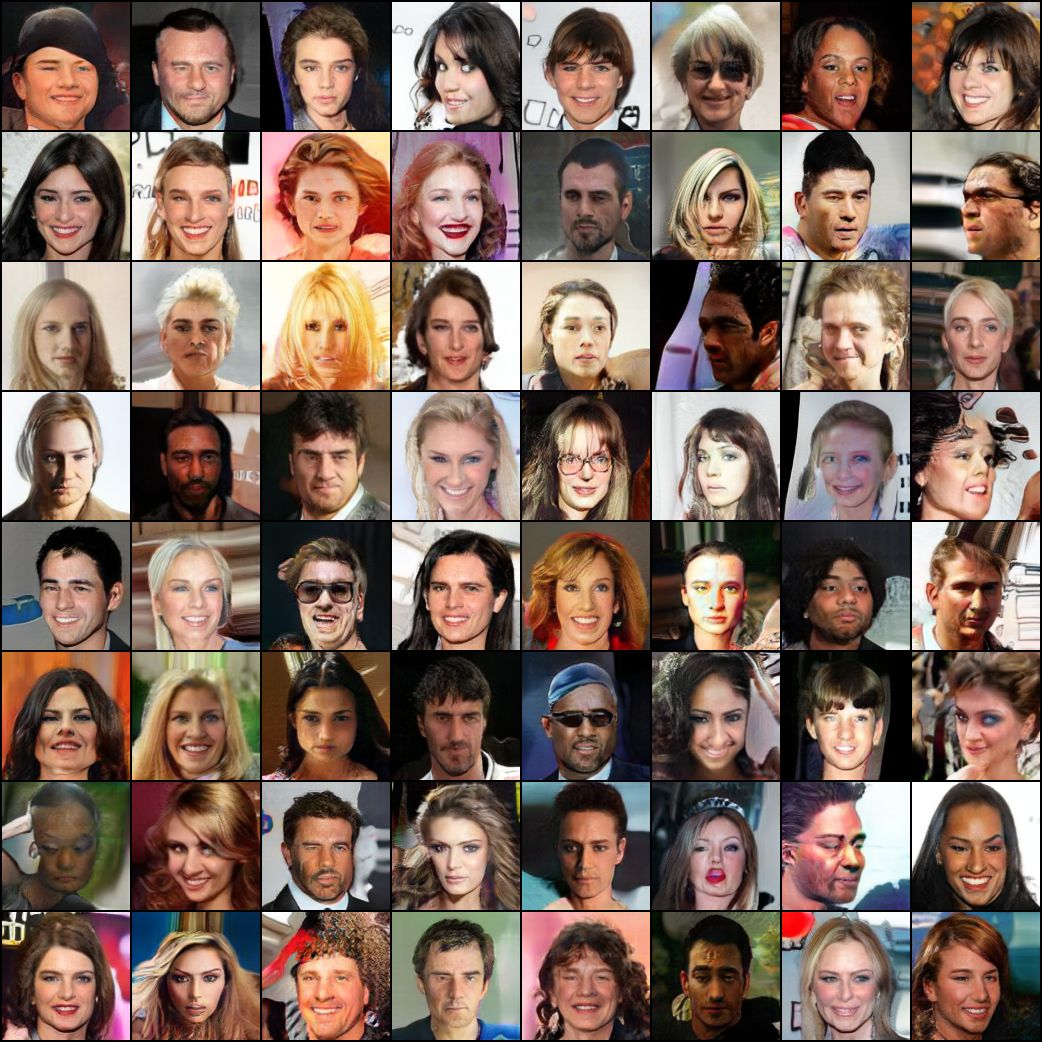}
        \caption{GraND-GAN}
    \end{subfigure}\hfill
        \begin{subfigure}[b]{0.33\textwidth}
        \includegraphics[width=1.0\textwidth]{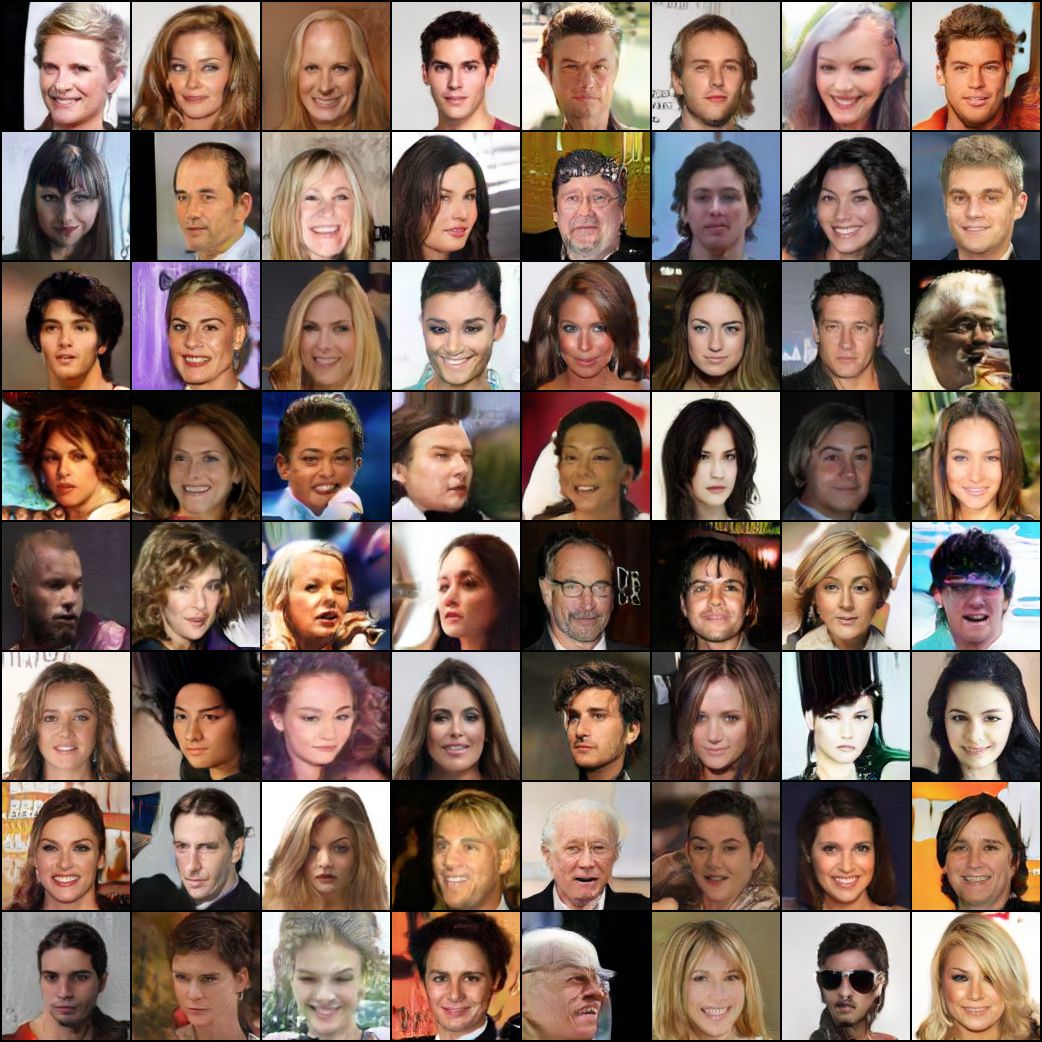}
         \caption{GraNC-GAN}
    \end{subfigure}
    \caption{\label{fig:qualitative-celeba} Qualitative results on CelebA across different models, including baselines (NSGAN, WGAN-GP, SNGAN), the best performing modified baseline (NSGAN-GP$\dagger$) and our methods (GraND-GAN and GraNC-GAN). Zoom in for better viewing.}
\end{figure*}

\begin{figure*}
	\centering 
    \begin{subfigure}[b]{0.33\textwidth}
		\includegraphics[width=1.0\textwidth]{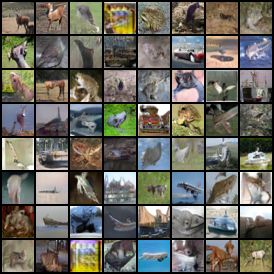}
        \caption{NSGAN}
    \end{subfigure} \hfill
    \begin{subfigure}[b]{0.33\textwidth}
		\includegraphics[width=1.0\textwidth]{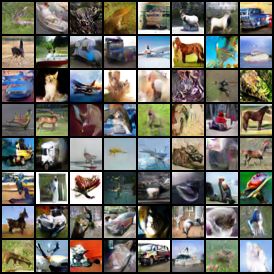}
        \caption{WGAN-GP}
    \end{subfigure}\hfill
    \begin{subfigure}[b]{0.33\textwidth}
		\includegraphics[width=1.0\textwidth]{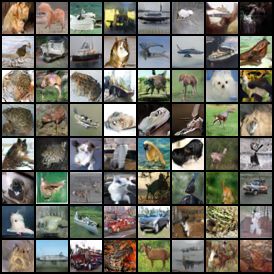}
        \caption{SNGAN}
    \end{subfigure}\\
    \begin{subfigure}[b]{0.33\textwidth}
		\includegraphics[width=1.0\textwidth]{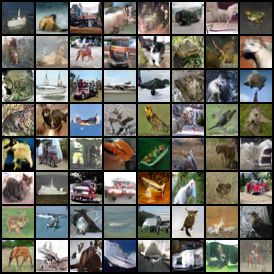}
        \caption{NSGAN-GP$\dagger$}
    \end{subfigure}\hfill
    \begin{subfigure}[b]{0.33\textwidth}
		\includegraphics[width=1.0\textwidth]{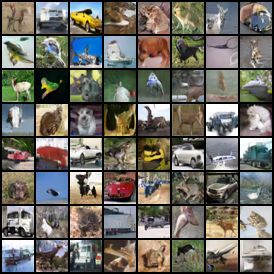}
        \caption{GraND-GAN}
    \end{subfigure}\hfill
        \begin{subfigure}[b]{0.33\textwidth}
		\includegraphics[width=1.0\textwidth]{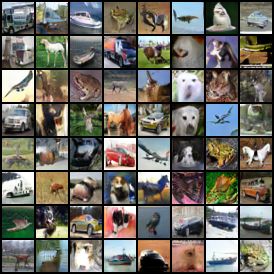}
        \caption{GraNC-GAN}
    \end{subfigure}
    \caption{\label{fig:qualitative-cifar10} Qualitative results on CIFAR-10 across different models, including baselines (NSGAN, WGAN-GP, SNGAN), the best performing modified baseline (NSGAN-GP$\dagger$) and our methods (GraND-GAN and GraNC-GAN). Zoom in for better viewing.}
\end{figure*}

\begin{figure*}
	\centering 
    \begin{subfigure}[b]{0.33\textwidth}
		\includegraphics[width=1.0\textwidth]{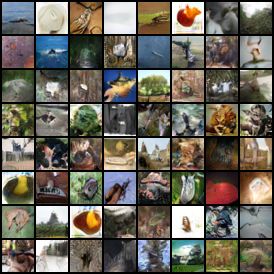}
        \caption{NSGAN}
    \end{subfigure} \hfill
    \begin{subfigure}[b]{0.33\textwidth}
		\includegraphics[width=1.0\textwidth]{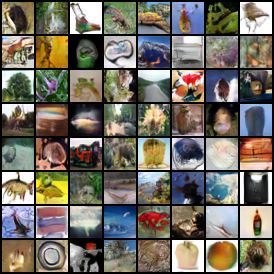}
        \caption{WGAN-GP}
    \end{subfigure}\hfill
    \begin{subfigure}[b]{0.33\textwidth}
		\includegraphics[width=1.0\textwidth]{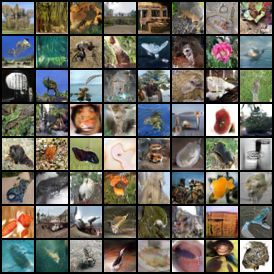}
        \caption{SNGAN}
    \end{subfigure}\\
    \begin{subfigure}[b]{0.33\textwidth}
		\includegraphics[width=1.0\textwidth]{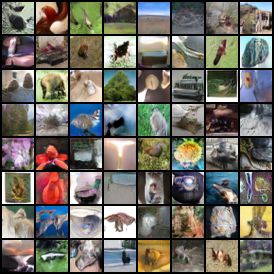}
        \caption{NSGAN-GP$\dagger$}
    \end{subfigure}\hfill
    \begin{subfigure}[b]{0.33\textwidth}
		\includegraphics[width=1.0\textwidth]{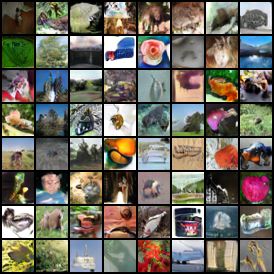}
        \caption{GraND-GAN}
    \end{subfigure}\hfill
        \begin{subfigure}[b]{0.33\textwidth}
		\includegraphics[width=1.0\textwidth]{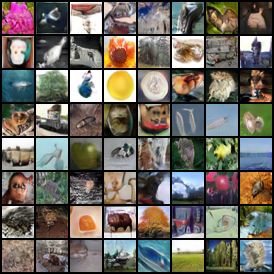}
        \caption{GraNC-GAN}
    \end{subfigure}
    \caption{\label{fig:qualitative-cifar100} Qualitative results on CIFAR-100 across different models, including baselines (NSGAN, WGAN-GP, SNGAN), the best performing modified baseline (NSGAN-GP$\dagger$) and our methods (GraND-GAN and GraNC-GAN). Zoom in for better viewing.}
\end{figure*}

\begin{figure*}
	\centering 
    \begin{subfigure}[b]{0.33\textwidth}
		\includegraphics[width=1.0\textwidth]{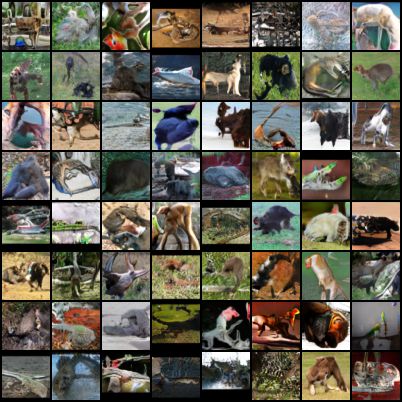}
        \caption{NSGAN}
    \end{subfigure} \hfill
    \begin{subfigure}[b]{0.33\textwidth}
		\includegraphics[width=1.0\textwidth]{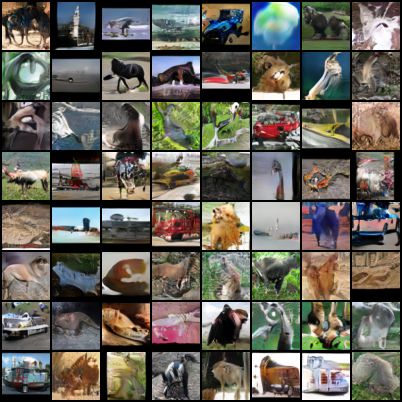}
        \caption{WGAN-GP}
    \end{subfigure}\hfill
    \begin{subfigure}[b]{0.33\textwidth}
		\includegraphics[width=1.0\textwidth]{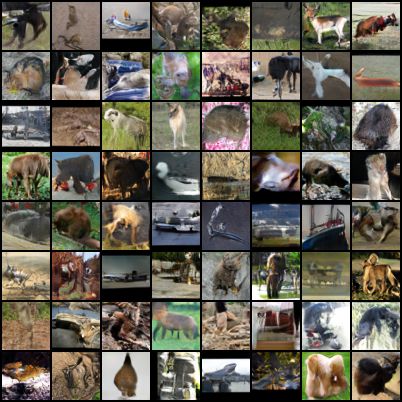}
        \caption{SNGAN}
    \end{subfigure}\\
    \begin{subfigure}[b]{0.33\textwidth}
		\includegraphics[width=1.0\textwidth]{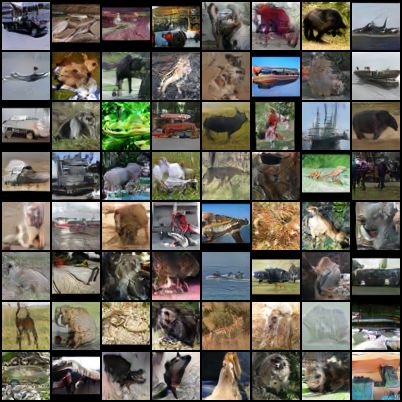}
        \caption{NSGAN-GP$\dagger$}
    \end{subfigure}\hfill
    \begin{subfigure}[b]{0.33\textwidth}
		\includegraphics[width=1.0\textwidth]{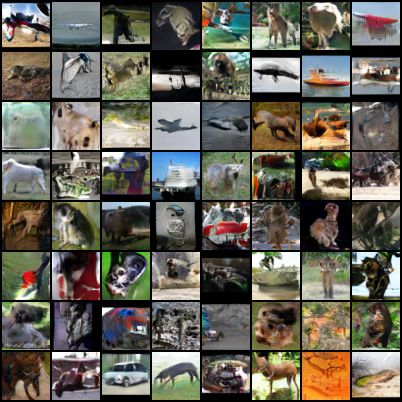}
        \caption{GraND-GAN}
    \end{subfigure}\hfill
        \begin{subfigure}[b]{0.33\textwidth}
		\includegraphics[width=1.0\textwidth]{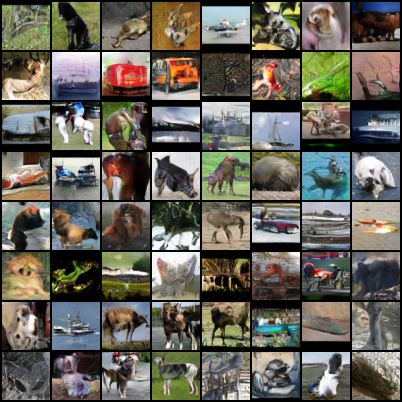}
        \caption{GraNC-GAN}
    \end{subfigure}
    \caption{\label{fig:qualitative-stl10} Qualitative results on STL-10 across different models, including baselines (NSGAN, WGAN-GP, SNGAN), the best performing modified baseline (NSGAN-GP$\dagger$) and our methods (GraND-GAN and GraNC-GAN). Zoom in for better viewing.}
\end{figure*}

Figures \ref{fig:qualitative-cifar10}, \ref{fig:qualitative-cifar100}, \ref{fig:qualitative-stl10}, \ref{fig:qualitative-lsun} and \ref{fig:qualitative-celeba} present a sample of images generated by different methods for CIFAR-10, CIFAR-100, STL-10, LSUN-Bedrooms and CelebA, respectively. We compare the results of our methods qualitatively with the baselines (NSGAN, WGAN-GP, SNGAN) and the best model of the modified baselines (NSGAN-GP$\dagger$ where $\dagger$ represents an adjusted Lipschitz constraint to match the piecewise Lipschitz constant of our methods).

\section{Soft versus Hard Hinge Performance}

We tested GraNC-GAN on both soft and hard hinge losses (recalling that the soft hinge loss is obtained by replacing the ReLU non-linearity in the standard hard hinge loss with the softplus activation).
On LSUN, GraNC-GAN struggles to converge with hard hinge, while it outperforms SNGAN with soft hinge loss.
Moreover, if one lowers the LRs on LSUN (to be those used by GraNC-GAN on CelebA; see \S\hyperref[appendixA]{A}), the soft hinge version performs better by ${\sim}$6 FID (specifically, 20.2 vs.\ 26.3).
On CelebA, using hard hinge resulted in an FID of 14, two points higher than that obtained via soft hinge (12), as displayed in the main paper.
Altogether, these suggest the soft hinge loss is generally more performant and stable than the standard hard hinge function, at least for GraN. Previous works, such as SNGAN, also note such instabilities across different loss functions, and, therefore, switch from the  Wasserstein loss to the (hard) hinge loss in their work. In our case, soft hinge loss was found to work the best.

\section{Effect of $\epsilon_{\text{Adam}}$ in the Adam update on GAN training}

To illustrate a qualitative effect of tuning $\epsilon_{\text{Adam}}$ in the Adam update on training GANs, we train GraNC-GAN on CIFAR-10 with Hinge loss for 1000 iterations, fixing the Lipschitz constant $\mathcal{K}=1$. We train two models, one with $\epsilon_{\text{Adam}}=1\times10^{-8}$ (default value) and another model with  $\epsilon_{\text{Adam}}=1\times10^{-7}$ (i.e., $10\times$ larger than the default). Figure \ref{fig:epsadam} show the qualitative results of a few examples sampled from the generators. 

As noted in the main paper, tuning the Lipschitz constant $\mathcal{K}$ has an effect that is equivalent to changing $\epsilon_{\text{Adam}}$. Figure \ref{fig:epsadam} qualitatively demonstrates that tuning $\epsilon_{\text{Adam}}$ (or $\mathcal{K}$, in effect) affects GAN training considerably.

\section{Stability of Modern GANs} 

Recent families of GANs, including those based on BigGAN \cite{brock2018large} and StyleGAN \cite{karras2019style}, have achieved unprecedented synthesis results; yet, they are not immune from instability issues. 
BigGAN devotes a significant portion of their paper to understanding stability (see, e.g., Sections 4.1 and 4.2 on ``characterizing instability"). Furthermore, they note that ``it is possible to enforce stability by strongly constraining D, but doing so incurs a dramatic cost in performance." 
Instability persists even within more recent methods that are based on BigGAN, such as U-net GAN \cite{schonfeld2020u}, which experiences ${\sim}40\%$ of its runs failing.
While StyleGAN does not present a stability analysis, their network relies heavily on progressive growing \cite{karras2017progressive} for stability, which induces artifacts (and additional training complexity) addressed in follow-up work (StyleGANv2 \cite{karras2020analyzing}). 
Similarly, MSG-GAN \cite{karnewar2020msg} demonstrates improved stability of its technique over progressive growing.
In other words, despite steady improvements, GAN stability remains a significant challenge, even for modern architectures.
See also
\cite{wiatrak2019stabilizing}
for a recent survey of stabilization techniques.

\end{document}